%% file: root.tex
\documentclass[journal]{IEEEtran}
\usepackage{amsmath,amsfonts}
\usepackage{algorithmic}
\usepackage{array}
\usepackage[caption=false,font=normalsize,labelfont=sf,textfont=sf]{subfig}
\usepackage{textcomp}
\usepackage{stfloats}
\usepackage{url}
\usepackage{verbatim}
\usepackage{graphicx}
\usepackage[noadjust]{cite}  

\usepackage{macros}
\usepackage[nolist]{acronym} 
\usepackage{bm}
\usepackage{amssymb}
\usepackage[dvipsnames]{xcolor}
\usepackage[ruled,vlined,linesnumbered]{algorithm2e}
\usepackage{microtype}
\usepackage{multirow}
\usepackage{cuted}
\usepackage{capt-of}
\usepackage{hyperref}

\newcommand{\algosection}[2][black]{%
  \BlankLine
  \noindent\textcolor{#1}{\hdashfill}\\[-0.4em]
  \textcolor{#1}{\tcp*[h]{#2}}\\[-0.4em]
  \noindent\textcolor{#1}{\hdashfill}
}
\newcommand{\hdashfill}{%
  \leavevmode
  \xleaders\hbox{-\,}\hfill\kern0pt
}

\definecolor{softil}{named}{MidnightBlue}
\definecolor{softpi}{named}{BrickRed}
\definecolor{softst}{named}{ForestGreen}
\definecolor{hardil}{named}{Plum}
\definecolor{hardvf}{named}{BlueViolet}

\begin{document}

\title{\textsc{ChainSplat}: A Physics-Inspired Screw-Theoretic Model for Learning Deformable Linear Object Dynamics from Multi-View RGB Videos}

\input{acronyms}

\author{Seungyeon Kim and No\'{e}mie Jaquier
\thanks{All authors are with the Department of Robotics, Perception and Learning, KTH Royal Institute of Technology. Emails: \href{mailto:seukim@kth.se}{\textrm{seukim@kth.se}}, \href{mailto:jaquier@kth.se}{\textrm{jaquier@kth.se}}. This work was supported by the Wallenberg Artificial Intelligence, Autonomous Systems, and Software Program (WASP) funded by the Knut and Alice Wallenberg Foundation and by the Swedish Research Council (Project: DefORM, Project ID 2025-05165\_VR). The computations were enabled by the Berzelius resource provided by the Knut and Alice Wallenberg Foundation and operated by NAISS.}
}


\maketitle

\begin{strip}
\vspace{-3cm}
    \centering
    \includegraphics[width=\linewidth]{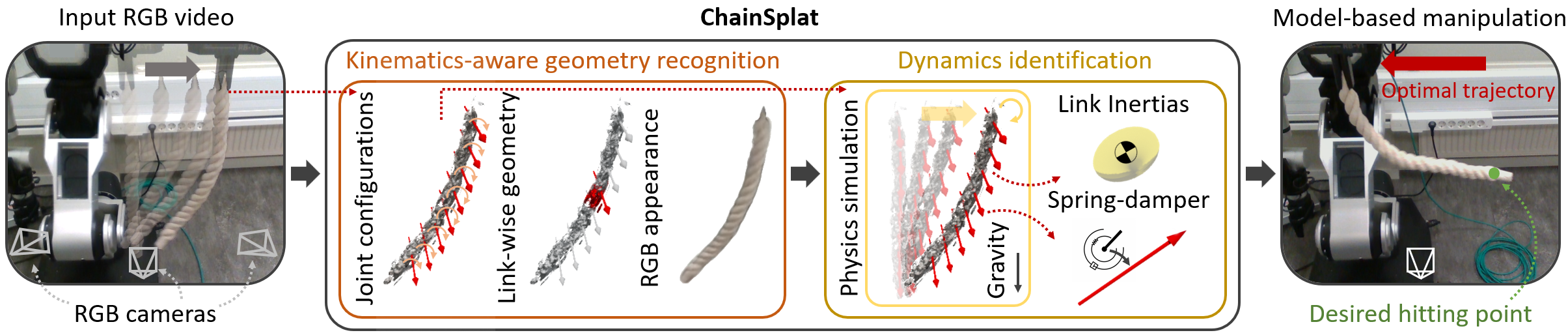}
    \captionof{figure}{\chainsplat overview. 
    (\textit{Kinematics-aware geometry recognition})  Given multi-view RGB videos of robot–object interactions, \chainsplat jointly estimates the joint configuration trajectory of a screw-based open-chain model and the link-wise geometry by minimizing the discrepancy between the observed and rendered RGB videos obtained via Gaussian Splatting.
    (\textit{Dynamics identification}) The dynamic parameters of \chainsplat are estimated by minimizing the discrepancy between the estimated and simulated joint trajectories obtained from physics simulation.
    (\textit{Model-based manipulation}) The optimized \chainsplat is used for model-based trajectory optimization for dynamic DLO manipulation tasks.
    }
    \label{fig:intro}
\end{strip}

\input{00abstract/v1_arxiv}
\input{01introduction/v6}
\input{02relatedworks/v4}
\input{03preliminaries/v5}
\input{04poesplat/v4}
\input{05optimization/v1}
\input{06manipulation/v3}
\input{07experiments/v1}
\input{08discussion/v2}
\input{09conclusion/v1}

\bibliographystyle{IEEEtran}
\bibliography{references}  

\end{document}

%% file: acronyms.tex
\newacro{ae}[AE]{Autoencoder}
\newacro{rom}[ROM]{reduced-order model}
\newacro{fom}[FOM]{full-order model}
\newacro{mor}[MOR]{model order reduction}
\newacro{lnn}[LNN]{Lagrangian neural network}
\newacro{hnn}[HNN]{Hamiltonian neural network}
\newacro{dof}[DoF]{degrees-of-freedom}
\newacro{ivp}[IVP]{initial value problem}
\newacro{dnn}[DNN]{deep neural network}
\newacro{spd}[SPD]{symmetric positive-definite}
\newacro{mlp}[MLP]{multilayer perceptron}
\newacro{fc}[FC]{fully-connected}
\newacro{gyrospd}[GyroSpd$_{\ty{++}}$]{gyrospace hyperplane-based}
\newacro{gyroai}[GyroAI]{gyrocalculus-based}
\newacro{rolnn}[RO-LNN]{reduced-order Lagrangian neural network}
\newacro{rohnn}[RO-HNN]{reduced-order Hamiltonian neural network}
\newacro{pde}[PDE]{partial differential equation}
\newacro{hnko}[HNKO]{Hamiltonian neural Koopman operator}
\newacro{mpc}[MPC]{model-predictive control}
\newacro{rl}[RL]{reinforcement learning}
\newacro{pd}[PD]{proportional derivative}
\newacro{iss}[ISS]{input-to-state stable}

%% file: 00abstract/v1_arxiv.tex

\begin{abstract}
Identifying the underlying dynamics and 3D geometry of deformable linear objects (DLOs), such as cables, ropes, and hoses, is essential for accurate robotic manipulation, but remains challenging due to their high-dimensional configuration spaces and diverse behaviors arising from varying material properties. 
Existing methods often rely on multi-stage pipelines and auxiliary depth inputs, which are prone to errors under dynamic interactions, while their high-dimensional state representations make model-based control computationally expensive.
In this paper, we introduce \chainsplat, a physics-inspired framework that jointly learns the 3D geometry, appearance, kinematics, and dynamics of DLOs solely from multi-view RGB videos. 
\chainsplat represents a DLO as an open-chain structure of rigid links connected by revolute joints, yielding an analytic, screw-theoretic model with a compact state representation parameterized by joint configurations. 
By integrating this formulation with Gaussian splatting, \chainsplat jointly recovers DLO dynamics, kinematics-aware 3D geometry, and appearance, while enabling high-fidelity RGB rendering from arbitrary states. 
Through real-world experiments, we demonstrate that \chainsplat achieves state-of-the-art performance in dynamics predictions, 3D geometry reconstruction, and RGB rendering across dynamic interactions. \chainsplat further enables real-time state and force estimation, as well as accurate model-based trajectory optimization, highlighting its practical utility for real-world robotic manipulation of DLOs.
Accompanying source code and video are
available at: \url{https://chainsplat.github.io}.
\end{abstract}


%% file: 01introduction/v6.tex
\section{Introduction}
Deformable linear objects (DLOs) --- flexible one-dimensional structures such as cables, ropes, and hoses --- are ubiquitous in everyday environments. Accurate robotic manipulation of such objects requires a precise understanding of their underlying physical dynamics~\cite{ai2025review}. 
Modeling these dynamics is particularly challenging due to the high-dimensional configuration space of deformable objects and the diverse behaviors arising from variations in material properties, which are often unknown a priori~\cite{arriola2020modeling, yin2021modeling}.

In this work, we address the challenging problem of inferring DLO dynamics directly from multi-view RGB videos of robot-object interactions. 
We argue that an effective state representation for dynamics modeling should satisfy two key criteria: (1) it should accurately capture the object's 3D geometry and appearance, and (2) it should be compact, yet expressive enough to provide computationally-efficient and accurate dynamics predictions. 
The former facilitates real-to-sim-to-real pipelines by constructing physically-realistic digital twins~\cite{jiang2025phystwin, chen2026empm}, thereby enabling flexible manipulation strategies (e.g., via reinforcement learning or model-based planning) and task-specific cost function design~\cite{torne2024reconciling}. The latter is essential for planning and control in real-world manipulation tasks~\cite{mamedov2024pseudo, mamedov2025learning} and real-time applications such as state estimation~\cite{xiang2023trackdlo, dinkel2025dlo}.

Data-driven approaches have recently attracted considerable attention for learning neural dynamics models from interaction data using visual state representations, demonstrating promising performance in modeling the dynamics of deformable objects~\cite{nair2017combining, yan2021learning, zhang2021deformable, ma2022learning, driess2023learning, liu2023model, zhang2024adaptigraph, zhang2025dynamic, duisterhof2024deformgs, huang2025particleformer, yan2020self, yang2021learning, yang2022learning, yu2022global, preiss2022tracking, caporali2024deformable}. 
However, these neural dynamics models do not explicitly enforce physical consistency, making them prone to overfitting to the training interaction distribution and thus limiting their ability to generalize to novel robot actions and unseen interaction scenarios.
Physics-inspired dynamics models address this limitation by incorporating physical laws as inductive biases, spanning a range of formulations from simple mass-spring particle systems~\cite{zhong2024reconstruction, jiang2025phystwin} to Lagrangian~\cite{friedl2025riemannian, friedl2026reduced}, Hamiltonian~\cite{friedl2025learning}, and continuum mechanics~\cite{zhang2024particle, chen2026empm}, as well as pseudo-rigid-body representations designed specifically for DLOs~\cite{mamedov2024pseudo, mamedov2025learning}. 
These approaches achieve improved accuracy in modeling deformable objects dynamics and exhibit stronger generalization to novel interactions. 

Several physics-inspired approaches explicitly capture an object's 3D geometry and appearance to construct physically-realistic digital twins from interaction videos~\cite{zhong2024reconstruction, zhang2024particle, jiang2025phystwin, chen2026empm}, closely aligning with our first criterion for an effective state representation. 
These approaches typically rely on multi-stage pipelines with intermediate components, including 2D point tracking~\cite{karaev2025cotracker3}, back-projection of depth images into 3D particles, and construction of 3D Gaussians whose motions are driven by the tracked particles for RGB rendering via Gaussian splatting~\cite{kerbl20233d}.
However, such pipelines not only increase system complexity, but also introduce multiple points of failure. 
In particular, 3D point trajectories are obtained by lifting 2D point tracks into 3D using depth observations, which are often noisier and less reliable than RGB images. This process is prone to errors, particularly for fast and dynamic motions, thus significantly degrading the overall performance of the pipeline. 
Moreover, by completely decoupling RGB rendering and dynamics learning, these approaches often exhibit limited RGB rendering fidelity. 
In addition, they typically represent the state of deformable objects via large sets of 3D particles, resulting in a high-dimensional state space and computationally-expensive dynamics models that make model-based control and state estimation prohibitively costly or even intractable. 
These limitations motivate not only the construction of more accurate digital twins, but also the design of compact state representation for efficient dynamics modeling, as per our second criterion.

\textbf{This paper introduces \chainsplat}, a DLO-specialized physics-inspired dynamics model that jointly learns the 3D geometry, appearance, kinematics, and dynamics of deformable linear objects. 
\chainsplat operates solely on RGB observations, eliminating the need for intermediate processing stages or auxiliary inputs such as depth images.
Its key idea is to represent a DLO as an open-chain structure in which a sequence of rigid links is connected by revolute joints, as shown in Figure~\ref{fig:intro} (orange box). 
This articulated representation admits an analytic kinematic model based on screw theory, enabling the efficient computation of each link's pose from joint configuration~\cite{lynch2017modern}. 
By assigning inertial properties and friction coefficients to each link together with spring-damper elements at each joint, we construct an analytic physics model~\cite{lynch2017modern, park2018geometric} that enables fast and physically-grounded simulation of DLO dynamics (see Figure~\ref{fig:intro}, yellow box).
This formulation naturally yields a low-dimensional dynamics model in which the system state is parameterized solely by the joint configuration of the open-chain structure.
We leverage Gaussian splatting to represent geometry and appearance~\cite{kerbl20233d}, further extending our open-chain formulation to model link-wise geometry and to render high-fidelity RGB images from arbitrary joint configurations.

We optimize the proposed \chainsplat from multi-view RGB videos, a task made challenging due to the tight coupling of the model's kinematics-aware 3D geometry, joint configurations, and dynamic parameters. 
To address this challenge, we formulate dynamics learning as a two-stage fully-differentiable optimization procedure, as shown in  Figure~\ref{fig:intro}-middle.
The first stage, termed kinematics-aware geometry recognition, jointly estimates the joint configuration trajectory of the open-chain model together with its link-wise geometry from multi-view RGB videos in an end-to-end manner. 
The second stage, referred as dynamics identification, then estimates the dynamic parameters of the \chainsplat model by minimizing the discrepancy between the simulated and estimated joint configuration trajectories.
As a result, \chainsplat accurately recovers the kinematics-aware geometry and appearance from input RGB videos, while yielding an efficient low-dimensional dynamics model of the DLO, in which the system state is parameterized by the joint configurations of the open-chain structure (see Figure~\ref{fig:intro}).

Building upon the \chainsplat model, we develop a simple yet effective framework for model-based DLO manipulation.
This framework leverages two key advantages of \chainsplat: its efficient low-dimensional dynamics model and its high-fidelity rendering capability from arbitratry states. 
The former leads to efficient model-based trajectory optimization for manipulation planning. In particular, as \chainsplat is fully differentiable, trajectory optimization can be performed at low computational cost via gradient-based methods. 
The latter enables real-time state estimation from a stream of RGB images by iteratively minimizing the discrepancy between the rendered and observed images, which can subsequently be used for state correction and closed-loop control.

Experiments on learning the dynamics of everyday DLOs show that \chainsplat achieves state-of-the-art performance in DLO digital twin modeling, leading to accurate dynamics modeling, faithful 3D geometry reconstruction, and high-quality RGB rendering, demonstrating that \chainsplat's open-chain structure provides effective inductive bias for DLO dynamics learning. 
We further demonstrate the effectiveness of our framework in real-time state estimation and manipulation, evaluated on the task of hitting a desired target point (see Figure~\ref{fig:intro}-right), highlighting its practical utility for real-world robotic manipulation of DLOs.

In summary, our contributions are threefold:
(1) We propose \chainsplat, a physics-inspired dynamics model that represents the geometry, appearance, kinematics, and dynamics of DLOs through a compact, physically-grounded open-chain representation based on screw theory (Section~\ref{sec:chainsplat});
(2) We develop an optimization framework that jointly recovers the geometry, kinematic states, and dynamic parameters of a \chainsplat model directly from multi-view RGB videos (Section~\ref{sec:chainsplat_optimization}); and
(3) We design efficient model-based state estimation and trajectory optimization frameworks built upon the optimized \chainsplat model, enabling effective real-world manipulation of DLOs (Section~\ref{sec:chainsplat_manipulation}).

%% file: 02relatedworks/v4.tex
\section{Related Works}
This section reviews existing approaches for learning deformable objects dynamics, with a particular focus on DLOs. 
We first discuss data-driven neural models and then review physics-inspired methods. 

\subsection{Learning Neural Dynamics of Deformable Linear Objects}
Early works attempted to directly learn deformable object dynamics from RGB videos using Convolutional Neural Networks (CNNs), modeling the dynamics in image space without explicit physical structure~\cite{nair2017combining, yan2021learning, zhang2021deformable}. 
To improve dynamics learning efficiency and representation quality, subsequent approaches extracted object-centric particles or keypoints and modeled their interactions explicitly. 
Representative examples include recurrent neural network (RNN)-based formulations predicting the 3D coordinates of the DLO endpoint~\cite{preiss2022tracking}, sparse neural network formulations operating on particle-based representations~\cite{liu2023model}, and Graph Neural Networks (GNNs), which capture relational dependencies between and propagate interaction effects across particles in deformable structures~\cite{ma2022learning, driess2023learning, zhang2024adaptigraph, zhang2025dynamic}.
Moreover, transformer-based architectures were recently explored for long-range modeling of deformable dynamics~\cite{huang2025particleformer}, while neural voxel encodings and multilayer perceptrons (MLPs) were employed as expressive models for handling highly deformable objects with shadows and occlusions~\cite{duisterhof2024deformgs}.

Several works designed dynamics models specifically tailored to DLOs~\cite{yan2020self, yang2021learning, yang2022learning, yu2022global, preiss2022tracking, caporali2024deformable}. 
Early works typically modeled DLOs as sequences of connected line segments and learned their dynamics in the coordinate space of their segment endpoints. 
For example, bi-directional LSTMs were used to predict dynamics in segment-based positional representations~\cite{yan2020self}. 
Later works combined LSTM architectures with GNNs to better capture both temporal and structural dependencies within deformable objects~\cite{yang2021learning, yang2022learning}. 
Other approaches employed lightweight MLPs together with physics-parameter-conditioned models to improve adaptability across material properties~\cite{caporali2024deformable}.
Beyond discrete segment-based representations, other methods modeled DLOs using continuous one-dimensional basis functions  (e.g., radial basis functions) and formulated Jacobian-based first-order differential dynamics models estimated through data-driven neural approximators~\cite{yu2022global}.
Although these neural dynamics models have shown promising performance in DLO dynamics prediction, they do not explicitly enforce physical consistency, thus limiting their ability to generalize beyond the interaction patterns observed during training.

\subsection{Physics-inspired Dynamics Models for Deformable Linear Objects}
Several recent works have explored physics-inspired dynamics models to improve the accuracy and generalizability of deformable object dynamics models under novel interactions. Representative efforts leveraged classical mechanics formulations to jointly learn autoencoder-based latent representations together with low-dimensional latent Lagrangian~\cite{friedl2025riemannian, friedl2026reduced} or Hamiltonian~\cite{friedl2025learning} dynamics. The resulting low-dimensional dynamics models have been successfully utilized for closed-loop control tasks~\cite{friedl2026reduced}.

Other works have incorporated explicit physical priors into dynamics models specialized for DLOs~\cite{chen2025differentiable, mamedov2024pseudo, mamedov2025learning}. 
Several approaches adopted physically-grounded energy-based formulations inspired by Discrete Elastic Rods (DER)~\cite{bergou2008discrete}, in which bending and twisting energies are explicitly modeled to derive the equations of motion while only residual dynamics are learned to improve generalization to novel interactions~\cite{chen2025differentiable}. 
Another line of work employs pseudo-rigid-body (PRB) representations~\cite{wittbrodt2006dynamics, moberg2014modeling}, which are conceptually more similar to \chainsplat. 
These methods parametrize DLOs using a small number of joint configurations and learn neural dynamics directly in the resulting low-dimensional state space~\cite{mamedov2024pseudo, mamedov2025learning}.
Both the latent-space-based and DER/PRB-based approaches have shown promising results for deformable object dynamics learning. They also aim to design generalizable and computationally-efficient dynamics models for control, sharing one of the key motivations of our work. However, they exclusively focus on dynamics learning. In contrast, our method recovers the geometry and appearance of DLOs while jointly learning an interpretable physics-based dynamics model directly from interaction videos.

Closer to our work, several recent approaches jointly learn physics-inspired dynamics together with the geometry and appearance of deformable objects from videos~\cite{jiang2025phystwin, zhang2024particle, chen2026empm}. 
Their pipeline consists of tracking a set of 2D points on deformable objects using CoTracker~\cite{karaev2025cotracker3}, back-projecting them onto 3D particles using depth images, constructing a set of 3D Gaussians by optimizing their appearance on the initial RGB frames using Gaussian splatting~\cite{kerbl20233d}, and rendering RGB videos by interpolating the motion of the 3D Gaussians from neighboring particle motions via Linear Blend Skinning (LBS)~\cite{sumner2007embedded}.
The aforementioned approaches primarily differ in their choice of dynamics models for the 3D particles.
PhysTwin~\cite{jiang2025phystwin} incorporates a mass-spring model~\cite{zhong2024reconstruction} to simulate 3D particle dynamics. 
To improve robustness and construct physically realistic digital twins from real-world observations, it augments the observed surface particles with interior particles that fill the object's volume using a 3D reconstruction module~\cite{xiang2025structured}. 
Particle-Grid Neural Dynamics (PGND)~\cite{zhang2024particle} proposes a hybrid Lagrangian-Eulerian neural dynamics model inspired by the Material Point Method (MPM), in which object particles and grid-based velocity fields are modeled separately. 
More recently, Embodied Material Point Method (EMPM)~\cite{chen2026empm} directly adopts an explicit MPM-based physics model for deformable object dynamics prediction. In contrast to PGND, EMPM identifies explicit material parameters, including physical properties such as Young's modulus.

Despite their enhanced performance in dynamics learning and generalization under novel interactions, these methods remain limited in several important aspects. 
First, they typically represent deformable objects using large sets of 3D particles, resulting in high-dimensional state representations that make planning and model-based control computationally expensive.
Second, their multi-stage pipelines increase complexity and lead to accumulated errors in 3D particle trajectory estimation, which are exacerbated by the noise inherent to depth observations. Moreover, decoupling the dynamics learning and RGB rendering limits the rendering fidelity.
In contrast, our approach directly learns a compact and efficient physics-based dynamics model of DLOs from multi-view RGB videos, without requiring intermediate processing stages or auxiliary depth observations.

%% file: 03preliminaries/v5.tex
\section{Preliminaries}
In this section, we introduce the three core components underlying our approach. 
We first briefly review the analytic kinematics and dynamics of open-chain systems.
We then introduce 3D Gaussian splatting, a representation and rendering framework that models 3D scenes using Gaussian primitives reconstructed from multi-view RGB images.

\subsection{Kinematics of Open-chain Systems}
\label{sec:kinematics_openchain}
Following~\cite{lynch2017modern}, we formulate the kinematics of open-chain systems using the special Euclidean group $\SEthree$ to represent rigid-body poses and screw theory to describe the relative motion between connected links. 

\subsubsection{Special Euclidean Group}
The special Euclidean group $\SEthree$ is the Lie group representing rigid-body motions in three-dimensional space. An element $\bm T \in \SEthree$ is given by
\begin{equation}
\bm T =
\left(
\begin{matrix}
\bm R & \bm p \\
0 & 1
\end{matrix}
\right)
\in \mathbb{R}^{4 \times 4},
\end{equation}
where $\bm R \in \SOthree$ is a rotation matrix and $\bm p \in \euclideanspace^3$ is a translation vector. For convenience, we equivalently write ${\bm T=(\bm R,\bm p)\in\SEthree}$.
The Lie algebra associated with $\SEthree$, denoted by $\sethree$, consists of elements of the form
\begin{equation}
\label{eq:lie_algebra_se3}
[\bm S] =
\left(
\begin{matrix}
[\bm \omega] & \bm v \\
0 & 0
\end{matrix}
\right)
\in \mathbb{R}^{4 \times 4},
\:\:\:
\bm S =
\left(
\begin{matrix}
\bm \omega \\
\bm v
\end{matrix}
\right)
\in \mathbb{R}^6,
\end{equation}
where $[\bm S]$ and $\bm S$ denote the matrix and vector representations of the same Lie algebra element, respectively. 
Here, $\bm\omega\in\mathbb{R}^3$ denotes the angular velocity component, whose corresponding skew-symmetric matrix representation is
\begin{equation}
[\bm \omega] =
\left(
\begin{matrix}
0 & -\omega_3 & \omega_2 \\
\omega_3 & 0 & -\omega_1 \\
-\omega_2 & \omega_1 & 0
\end{matrix}
\right)
\in \mathbb{R}^{3 \times 3},
\:\:\:
\bm \omega =
\left(
\begin{matrix}
\omega_1 \\
\omega_2 \\
\omega_3
\end{matrix}
\right)
\in \mathbb{R}^3,
\end{equation}
and $\bm v\in\mathbb{R}^3$ denotes the linear velocity component. 
We also write $\bm S=(\bm\omega,\bm v)\in\sethree$, which is commonly referred to as a {\it twist}.
Throughout this paper, elements of $\sethree$ are used in two contexts, namely: (1) to parameterize the joint motions of open-chain systems within the screw theory framework, and (2) to represent the spatial velocities of individual links.

The dual space of the Lie algebra $\sethree$, denoted by $\sethree^*$, is a $6$-dimensional vector space composed of linear functionals on $\sethree$. An element of $\sethree^*$ is typically written as $\bm F=(\bm\tau,\bm f)$, referred to as a {\it wrench}, and represents a spatial force. Intuitively, a wrench maps a twist $\bm V=(\bm\omega,\bm v)\in\sethree$ to the scalar power $W = \bm \tau \cdot \bm \omega + \bm f \cdot \bm v$. Wrenches play a crucial role in describing the physical dynamics introduced in Section~\ref{subsec:dynamics_open_chain}.

\subsubsection{Screw Theory}
Screw theory provides a natural mathematical framework for describing rigid-body motions of robotic manipulators, particularly for open-chain systems composed of revolute and prismatic joints. 
For a given reference frame, a screw axis is represented by an element ${\bm S=(\bm\omega,\bm v)\in\sethree}$. 
For a revolute joint, $\bm\omega\in\mathbb{R}^3$ is a unit vector specifying the joint's axis of rotation, while the linear component is given by $\bm v=-\bm\omega\times\bm q$, where $\bm q\in\mathbb{R}^3$ denotes an arbitrary point on the screw axis. 
For a prismatic joint, the screw axis represents pure translation along the direction $\bm v$ with $\|\bm v\|=1$ and $\bm\omega=\bm0$. 
Throughout this work, we model open-chain system composed exclusively of revolute joints.

\begin{figure}[!t]
    \centering
    \includegraphics[width=\linewidth]{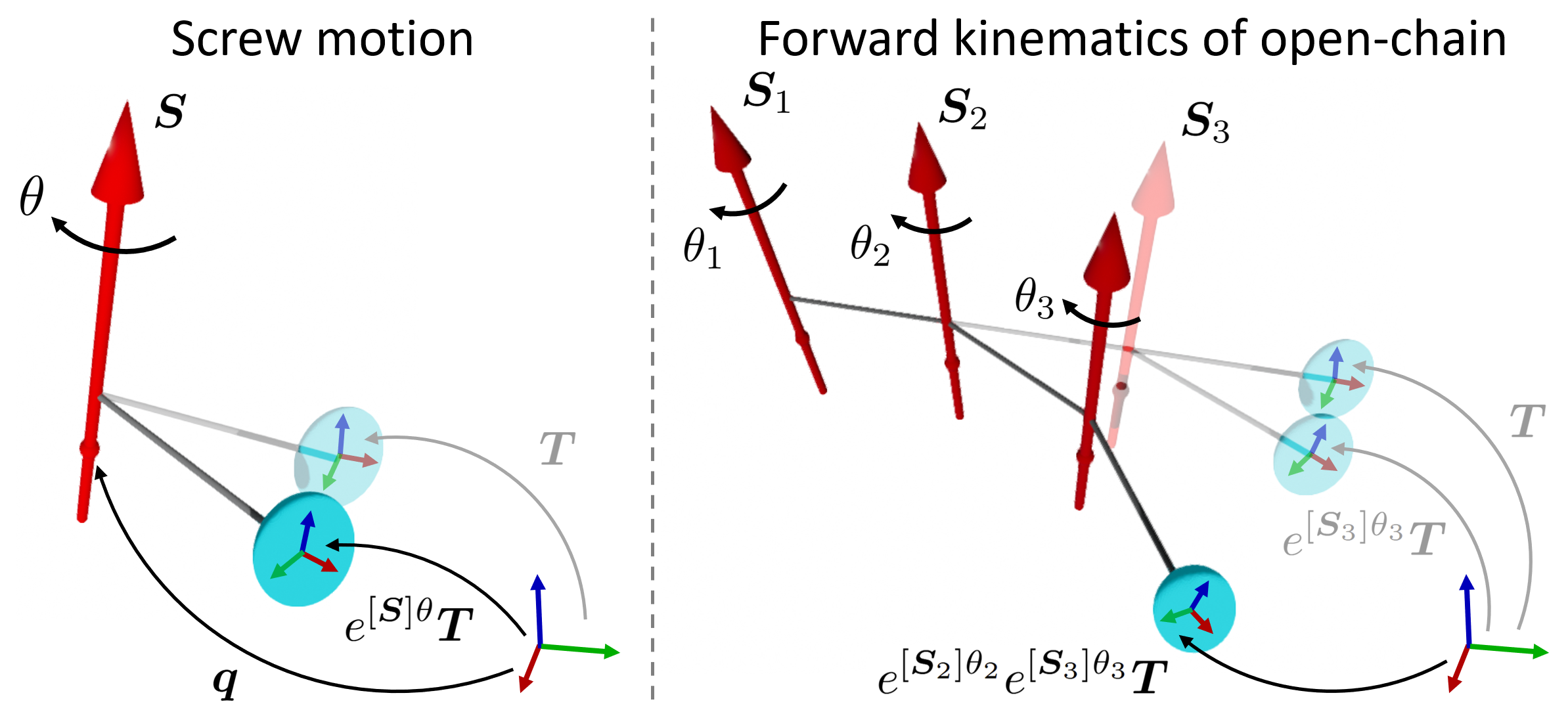}
    \vspace{-15pt}
    \caption{Illustration of screw motion and forward kinematics of an open-chain system. 
    (\textit{Screw motion}) Given a screw axis $\bm S$ and a joint displacement $\theta$, the blue ellipsoid with pose $\bm T$ is transformed by the matrix exponential.
    (\textit{Forward kinematics of open-chain}) The pose $\bm T$ of the blue ellipsoid transformed by the last screw motion is given by $e^{[\bm S_3]\theta_3}\bm T$. Considering the preceding screw motion similarly introduces another matrix exponential on the left. Repeating this process along the open chain yields the PoE formula~\eqref{eq:PoE-formulation}.
    }
    \vspace{-15pt}
    \label{fig:screw_and_poe}
\end{figure}

Given a screw axis $\bm S$ and a joint displacement $\theta \in \mathbb{R}$, a rigid body configuration $\bm T \in \SEthree$ is transformed along the screw axis following the matrix exponential as
\begin{equation}
\bm T' = e^{[\bm S]\theta} \bm T \in \SEthree,
\end{equation}
where $[\bm S]$ is the matrix representation of the screw axis $\bm S$ as in~\eqref{eq:lie_algebra_se3}, see  Figure~\ref{fig:screw_and_poe}-left. 
Closed-form expressions for the matrix exponential and further details on screw theory can be found in~\cite[Ch.~3]{lynch2017modern}.

\subsubsection{Forward Kinematics of Open-Chain Systems}
Consider an open-chain system consisting of $n$ rigid links connected serially by $n$ joints, whose base and end-effector frames are attached to the fixed base and the terminal link, respectively. 
The forward kinematics describes the rigid-body configuration $\bm T \in \SEthree$ of the end-effector frame relative to the base frame as a function of the joint configuration ${\bm \theta = (\theta_1, \ldots, \theta_n) \in \mathbb{R}^n}$. 
Using the product-of-exponentials (PoE) formula~\cite[Ch.~4]{lynch2017modern}, the forward kinematics is given by
\begin{equation}
\label{eq:PoE-formulation}
\bm T(\bm \theta)
=
e^{[\bm S_1]\theta_1}
e^{[\bm S_2]\theta_2}
\cdots
e^{[\bm S_n]\theta_n}
\bm T(\bm 0),
\end{equation}
where $\bm T(\bm 0) \in \SEthree$ denotes the home configuration of the end-effector frame when $\bm\theta=\bm0$, and $\bm S_i\in\sethree$ denotes the screw axis of joint $i\in\{1,\ldots,n\}$ expressed in the base frame at the home configuration, as shown in Figure~\ref{fig:screw_and_poe}-right. 
More generally, the rigid-body configuration of a frame attached to the $j$-th link is given by
\begin{equation}
\bm T_j(\bm \theta)
=
e^{[\bm S_1]\theta_1}
e^{[\bm S_2]\theta_2}
\cdots
e^{[\bm S_j]\theta_j}
\bm T_j(\bm 0),
\label{eq:poe_formula}
\end{equation}
where $\bm T_j(\bm 0) \in \SEthree$ denotes the home configuration of the frame attached to the $j$-th link. This formulation provides a natural way to compute the transformed poses of each link of a DLO modeled as an open-chain system, as well as the Gaussian primitives associated with its geometry in \chainsplat.

\subsection{Dynamics of Open-chain Systems}
\label{subsec:dynamics_open_chain}
Next, we formulate the dynamics of open-chain systems based on the kinematic model introduced in Section~\ref{sec:kinematics_openchain} and on the $\SEthree$ formulation of single rigid-body dynamics with adjoint mappings following~\cite{lynch2017modern}.

\subsubsection{Large and Small Adjoint Mappings}
Here, we describe the large and small adjoint mappings, both of which act on elements of the Lie algebra (i.e., twists) but characterize transformations at the Lie group and Lie algebra levels, respectively. The small adjoint is obtained by differentiating the large adjoint.

Consider a rigid body transformation $\bm T = (\bm R, \bm p) \in \SEthree$ and a twist $\bm S = (\bm \omega, \bm v) \in \sethree$. The large adjoint mapping ${\Adjoint_{\bm T} : \sethree \rightarrow \sethree}$ describes the transformation of the coordinates of the twist $\bm S$ under the rigid-body transformation $\bm T$ and is defined as
$[\Adjoint_{\bm T}(\bm S)]
=
\bm T [\bm S] \bm T^{-1}$. 
Its vector form is given by
\begin{equation}
\Adjoint_{\bm T}(\bm S)
=
[\Adjoint_{\bm T}] \bm S
=
\left(
\begin{matrix}
\bm R & \bm 0 \\
[\bm p]\bm R & \bm R
\end{matrix}
\right)
\left(
\begin{matrix}
\bm \omega \\
\bm v
\end{matrix}
\right),
\end{equation}
where $[\Adjoint_{\bm T}] \in \mathbb{R}^{6 \times 6}$ denotes the matrix representation of the large adjoint operator.
Throughout this work, the large adjoint mapping is used to describe the coordinate transformations of spatial velocities and applied forces between link coordinate frames.

Given two twists $\bm S = (\bm \omega_S, \bm v_S)$, ${\bm V = (\bm \omega_V, \bm v_V) \in \sethree}$, the small adjoint mapping $\adjoint_{\bm S} : \sethree \rightarrow \sethree$ describes the infinitesimal change of the twist $\bm V$ induced by the motion generated by $\bm S$. It is defined as
$
[\adjoint_{\bm S}(\bm V)]
=
[\bm S][\bm V] - [\bm V][\bm S]$,
with corresponding vector form
\begin{equation}
\adjoint_{\bm S}(\bm V)
=
[\adjoint_{\bm S}] \bm V
=
\left(
\begin{matrix}
[\bm \omega_S] & \bm 0 \\
[\bm v_S] & [\bm \omega_S]
\end{matrix}
\right)
\left(
\begin{matrix}
\bm \omega_V \\
\bm v_V
\end{matrix}
\right),
\end{equation}
where $[\adjoint_{\bm S}] \in \mathbb{R}^{6 \times 6}$ denotes the matrix representation of the small adjoint operator. The small adjoint mapping is used to describe the Coriolis and centrifugal coupling terms of spatial rigid-body dynamics.

\subsubsection{Single Rigid-Body Dynamics}
\label{subsecsec:single_dynamics}
Consider a rigid body with a body-fixed reference frame $\{b\}$ attached to its center of mass. Let $\bm V_b = (\bm \omega_b, \bm v_b) \in \sethree$ denote the rigid body's spatial velocity expressed in frame $\{b\}$. 
When the reference frame follows a trajectory $\bm T_b(t)\in\SEthree$, its spatial velocity is given by
\begin{equation}
\bm V_b(t)
=
\log
\left(
\bm T_b(t)^{-1}\dot{\bm T}_b(t)
\right)
\in \sethree.
\label{eq:spatial_velocity}
\end{equation}
We denote the mass and the inertia matrix about the rigid body's center of mass as $m_b \in \mathbb{R}_+$ and $\bm I_b \in \mathbb{R}^{3\times3}$, respectively, and define the spatial inertia matrix as
\begin{equation}
\bm G_b =
\left(
\begin{matrix}
\bm I_b & \bm 0 \\
\bm 0 & m_b \bm I_3
\end{matrix}
\right)
\in \mathbb{R}^{6 \times 6}
\label{eq:spatial_inertia}
\end{equation}
where $\bm I_3 \in \mathbb{R}^{3 \times 3}$ denotes the identity matrix. 

Let $\bm F_b=(\bm \tau_b,\bm f_b)\in\sethree^*$ denote the externally applied wrench, where $\bm\tau_b \in\mathbb{R}^3$ and $\bm f_b\in\mathbb{R}^3$ denote the torque and force components expressed in frame $\{b\}$, respectively. The rigid-body dynamics can then be written as
\begin{equation}
\bm F_b 
=
\bm G_b \dot{\bm V}_b
-
\adjoint_{\bm V_b}^\trsp (\bm G_b \bm V_b) =
\bm G_b \dot{\bm V}_b
-
[\adjoint_{\bm V_b}]^\trsp \bm G_b \bm V_b. 
\end{equation}
We note that, for another reference frame $\{a\}$ attached to the body, the dynamics equations are expressed as
\begin{equation}
\bm F_a 
=
\bm G_a \dot{\bm V}_a
-
[\adjoint_{\bm V_a}]^\trsp \bm G_a \bm V_a,
\end{equation}
where $\bm F_a$ and $\bm V_a$ are the wrench and spatial velocity expressed in frame $\{a\}$, respectively. They are related to the corresponding quantities expressed in frame $\{b\}$ by
\begin{equation}
\bm V_a
=
[\Adjoint_{\bm T_{ab}}]
\bm V_b,
\:\:\:
\bm F_a
=
[\Adjoint_{\bm T_{ba}}]^\trsp
\bm F_b,
\end{equation}
while the spatial inertia matrix $\bm G_a$ relates to $\bm G_b$ as
\begin{equation}
\bm G_a 
=
[\Adjoint_{\bm T_{ba}}]^\trsp \bm G_b [\Adjoint_{\bm T_{ba}}].
\label{eq:spatial_inertia_transform}
\end{equation}

\begin{figure}
    \centering
    \includegraphics[width=\linewidth]{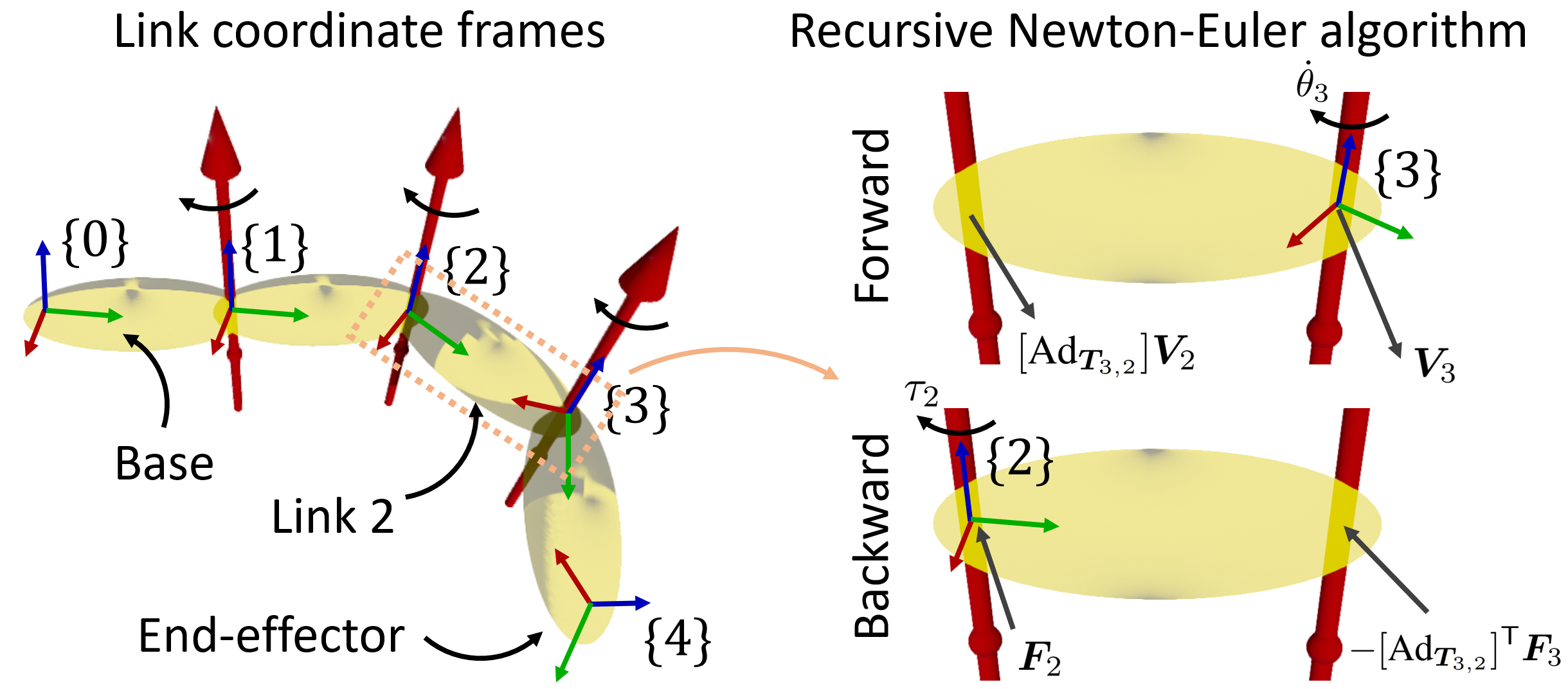}
    \vspace{-15pt}
    \caption{Illustration of link coordinate frames and recursive Newton--Euler algorithm. 
    (\textit{Link coordinate frames}) Each link frame $\{i\}$ is attached to the joint $i$, while the base frame $\{0\}$ and the end-effector frame $\{n+1\}$ (here, $n=3$) are attached to the fixed base and the end-effector, respectively.
    (\textit{Newton--Euler algorithm}) The forward iteration~\eqref{eq:newton_euler_forward_1}--\eqref{eq:newton_euler_forward_2} and backward iteration~\eqref{eq:newton_euler_backward}--\eqref{eq:newton_euler_backward_2} are expressed in frames $\{i+1\}$ and $\{i\}$ (here, $i=2$), respectively.
    }
    \vspace{-15pt}
    \label{fig:recursive_dynamics}
\end{figure}

\subsubsection{Lie Group-based Dynamics of Open-chains}
\label{sec:dynamics_open_chain}
Next, we formulate the dynamics of open-chain systems based on the dynamics of single rigid bodies. 
Consider an open-chain system composed of $n$ rigid links connected serially by $n$ joints with screw axis $\bm S_i\in\sethree$. 
We first attach the link frame $\{i\}$ to joint $i$, the base frame $\{0\}$ to the fixed base, and the end-effector frame $\{n+1\}$ to the end-effector, as illustrated in Figure~\ref{fig:recursive_dynamics}-left. Expressed in frame $\{i\}$, the spatial inertia matrix of link $\{i\}$ is $\bm G_i\in\mathbb{R}^{6\times6}$ (see~\eqref{eq:spatial_inertia} and~\eqref{eq:spatial_inertia_transform}) and the screw axis is $\bm A_i = [\Adjoint_{T_{i,0}}] \bm S_i\in\sethree$.

Let $\bm V_i \in \sethree$ and $\bm F_i \in \sethree^*$ denote the spatial velocity link frame $i$ and spatial wrench transmitted to link $i$ through joint $i$, respectively, see Figure~\ref{fig:recursive_dynamics}-right. 
Assuming a fixed base, the base spatial velocity is $\bm V_0 = \bm 0$ and the base spatial acceleration is usually set to $\dot{\bm V}_0 = (\bm \omega_0, \bm v_0) = (0, -\bm g)$, where $\bm g \in \mathbb{R}^3$ is the gravity acceleration vector. The wrench $\bm F_{n+1}$ is exerted by the end-effector frame $\{n+1\}$ on the environment.

Under this setting, we formulate the open-chain dynamics using the recursive Newton-Euler algorithm~\cite{lynch2017modern,park2018geometric}, which consists of a forward iteration followed by a backward iteration. 
During the forward iteration, the positions, velocities, and accelerations of the links are propagated from the base to the end link. 
Specifically, given the joint positions $\bm\theta$, velocities $\dot{\bm\theta}$, and accelerations $\ddot{\bm\theta}$, we recursively compute
\begin{align}
\bm V_i
& =
[\Adjoint_{\bm T_{i, i-1}}]
\bm V_{i-1}
+
\bm A_i \dot{\theta}_i, 
\label{eq:newton_euler_forward_1}
\\
\dot{\bm V}_i
& =
[\Adjoint_{\bm T_{i, i-1}}]
\dot{\bm V}_{i-1}
+
[\adjoint_{\bm V_i}]\bm A_i\dot{\theta}_i
+
\bm A_i \ddot{\theta}_i.
\label{eq:newton_euler_forward_2}
\end{align}
for $i=1$ to $n$.
During the backward iteration, the forces and moments acting on the links are propagated from the end link to the base according to the single rigid-body dynamics formulation presented in Section~\ref{subsecsec:single_dynamics}.
Specifically, we recursively compute, for $i=n$ to $1$,
\begin{align}
\bm F_i
& =
[\Adjoint_{\bm T_{i+1, i}}]^{\trsp}
\bm F_{i+1}
+
\bm G_i \dot{\bm V}_i
-
[\adjoint_{\bm V_i}]^{\trsp}\bm G_i \bm V_i
\label{eq:newton_euler_backward}
\\ 
\tau_i
& =
\bm F_i^\top \bm A_i.
\label{eq:newton_euler_backward_2}
\end{align}
The above recursive formulation~\eqref{eq:newton_euler_forward_1}-\eqref{eq:newton_euler_backward_2} yields the inverse dynamics of the open-chain system. The equations of motion are equivalently expressed in the closed-form standard manipulator form as
\begin{equation}
\bm \tau
=
\bm M(\bm \theta)\ddot{\bm \theta}
+
\bm C(\bm \theta, \dot{\bm \theta})\dot{\bm \theta}
+
\bm g(\bm \theta)
+
\bm J(\bm \theta)^\top \bm F_{n+1},
\label{eq:lie_group_dynamics}
\end{equation}
where $\bm M(\bm\theta)\in\mathbb{R}^{n\times n}$ is the mass-inertia matrix, ${\bm C(\bm\theta,\dot{\bm\theta})\in\mathbb{R}^{n\times n}}$ represents the Coriolis and centrifugal terms, $\bm g(\bm\theta)\in\mathbb{R}^n$ is the gravity vector, and $\bm J(\bm\theta)$ is the manipulator Jacobian of the end-effector frame $\{n+1\}$, at which the wrench $\bm F_{n+1}$ is applied.
These quantities are expressed in closed form as
\begin{align}
\bm M(\bm \theta)
& =
\bm A^\top
\bm L(\bm \theta)^\top
\bm G
\bm L(\bm \theta)
\bm A,
\label{eq:inverse_dynamics_mass}
\\
\bm C(\bm \theta, \dot{\bm \theta})
& =
-\bm A^\top
\bm L(\bm \theta)^\top
\left(
\bm G
\bm L(\bm \theta)
[\adjoint_{\bm A \dot{\bm \theta}}]
+
[\adjoint_{\bm V}]^{\trsp}
\bm G
\right)
\bm L(\bm \theta)
\bm A, 
\label{eq:inverse_dynamics_coriolis}
\\
\bm g(\bm \theta)
& =
\bm A^\top
\bm L(\bm \theta)^\top
\bm G
\bm L(\bm \theta)
\dot{\bar{\bm V}}_{\mathrm{0}},
\label{eq:inverse_dynamics_gravity}
\end{align}
where the quantities without link indices denote the corresponding stacked quantities, e.g.,
\begin{align}
\bm V
&=
\left(\bm V_1, \ldots, \bm V_n\right)
\in \mathbb{R}^{6n}, 
\label{eq:stacked_velocity}
\\
\bm A
&=
\operatorname{diag}\left(\bm A_1, \ldots, \bm A_n\right)
\in \mathbb{R}^{6n\times n}, \\
[\adjoint_{\bm V}]
&=
\operatorname{diag}\left(
[\adjoint_{\bm V_1}], \ldots, [\adjoint_{\bm V_n}]
\right)
\in \mathbb{R}^{6n\times 6n}.
\end{align}
In addition, we define $\bar{\bm V}_0$ as
\begin{equation}
\bar{\bm V}_0
=
\left(
[\Adjoint_{\bm T_{10}}]\bm V_0,\,
\bm 0_{6(n-1)}
\right)
\in
\mathbb{R}^{6n},
\label{eq:bar_operation}
\end{equation}
which is introduced to obtain a compact closed-form expression for the gravity term.
Moreover, $\bm L(\bm\theta)$ is a block lower-triangular matrix composed of large adjoint operators that describe the kinematic propagation along the articulated chain. Detailed derivations can be found in~\cite[Ch.~8]{lynch2017modern}. This open-chain dynamics formulation serves as the foundation of the DLO dynamics model in \chainsplat, where it is extended to account for robot-object interactions.

\subsection{3D Gaussian Splatting}
\label{sec:3dgs}
3D Gaussian splatting is a differentiable rendering framework originally developed for novel-view synthesis from multi-view RGB images, which has also been adopted for 3D scene representation~\cite{kerbl20233d}.
Gaussian splatting represents a scene as a collection of 3D Gaussian primitives, where the $i$-th Gaussian $\mathcal{G}_i$ is parameterized by the tuple 
\begin{equation}
    \mathcal{G}_i =(\bm T_i, \bm s_i, \sigma_i, \bm c_i).
    \label{eq:gaussian_splatting_components}
\end{equation}
The frame ${\bm T_i = (\bm R_i, \bm \mu_i) \in \SEthree}$ denotes the pose of the Gaussian with position $\bm \mu_i \in \mathbb{R}^3$ and orientation $\bm R_i \in \SOthree$,  the parameter $\bm s_i \in \mathbb{R}_+^3$ defines the anisotropic scale of the Gaussian whose covariance matrix is given by
\begin{equation}
\bm \Sigma_i
=
\bm R_i
\mathrm{diag}(\bm s_i)^2
\bm R_i^\top,
\end{equation}
the scalar $\sigma_i \in [0,1]$ denotes the opacity of the Gaussian, and $\bm c_i \in \mathbb{R}^{3(L+1)^2}$ represents the surface color of the Gaussian ellipsoid, parameterized by spherical harmonics coefficients of degree $L$.

The colored Gaussians are then used to render an RGB image through $\alpha$-blending.  
Given a set $\mathcal{O}$ of depth-ordered Gaussians overlapping a pixel, the rendered pixel color $\bm C$ is computed using $\alpha$-blending as
\begin{equation}
\bm C
=
\sum_{i \in \mathcal{O}}
\bm c_i \alpha_i
\prod_{j=1}^{i-1}
(1-\alpha_j),
\label{eq:alpha_blending}
\end{equation}
where $\alpha_i$ is the scaled Gaussian function of the $i$-th Gaussian in 3D space defined as
\begin{equation}
\alpha_i(\bm x)
=
\sigma_i
\exp
\left(
-\frac{1}{2}
(\bm x - \bm \mu_i)^\top
\bm \Sigma_i^{-1}
(\bm x - \bm \mu_i)
\right),
\label{eq:alpha_function}
\end{equation}
where $\bm x \in \mathbb{R}^3$.
The Gaussian parameters are optimized by minimizing the rendering objective
\begin{equation}
\label{eq:rendering_loss}
\mathcal{L}_{\mathrm{render}}
=
(1-\lambda)\mathcal{L}_1
+
\lambda \mathcal{L}_{\mathrm{D\text{-}SSIM}},
\end{equation}
where $\mathcal{L}_1$ and $\mathcal{L}_{\mathrm{D\text{-}SSIM}}$ denote the pixel-wise $\ell_1$ loss and the D-SSIM perceptual loss between the rendered and ground-truth RGB images, respectively. Following prior work~\cite{kerbl20233d}, the weighting parameter $\lambda$ is typically set to $0.2$.
In this work, we adopt Gaussian primitives to represent the link-wise geometry of DLOs and employ $\alpha$-blending~\eqref{eq:alpha_function}-\eqref{eq:alpha_blending} together with the rendering loss functions~\eqref{eq:rendering_loss} to optimize \chainsplat models from multi-view RGB videos.

%% file: 04poesplat/v4.tex
\section{ChainSplat: Integrating Lie Group Dynamics with 3D Gaussians}
\label{sec:chainsplat}
This section introduces \chainsplat, a physics-inspired open-chain formulation for modeling the geometry, appearance, kinematics, and dynamics of DLOs. 
First, Section~\ref{subsec:ChainSplat_components} describes \chainsplat's core components that compose its underlying open-chain structure associated with a set of 3D Gaussians to capture the DLO geometry and appearance. 
Then, Section~\ref{subsec:ChainSplat_rgb_render} develops a differentiable RGB rendering function that renders \chainsplat from a given joint configuration, enabling fully differentiable optimization of the geometry, appearance, and joint configurations from multi-view RGB videos.
Finally, Section~\ref{subsec:ChainSplat_dynamics} introduces the corresponding dynamics formulation based on screw theory for physics-based simulation of DLOs.
The optimization of \chainsplat's parameters based on synchronized multi-view RGB videos of robot--DLO interactions is then described in Section~\ref{sec:chainsplat_optimization}. 

\begin{figure*}[tbp]
    \centering
    \includegraphics[width=\linewidth]{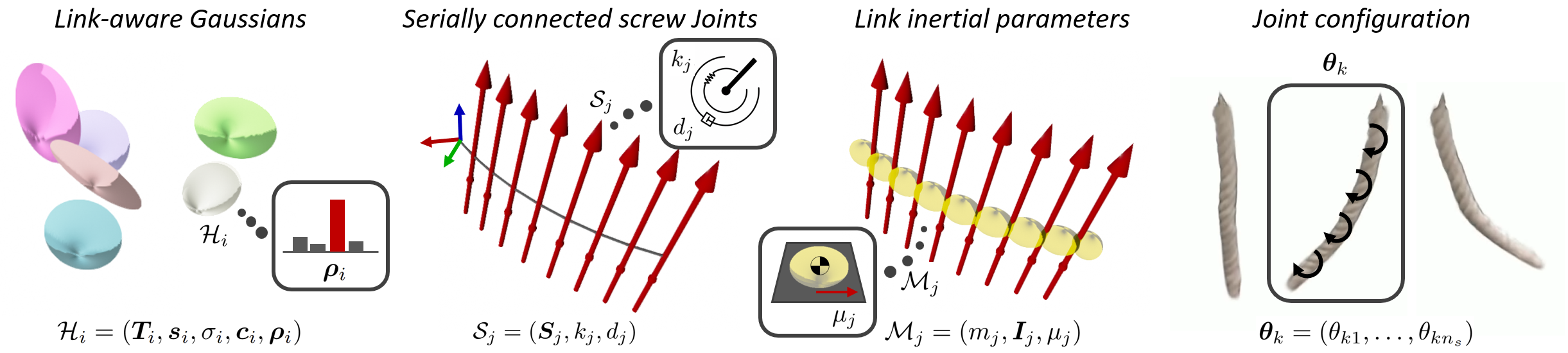}
    \vspace{-15pt}
    \caption{Core components of \chainsplat. 
    (\textit{Link-aware Gaussians}) Each Gaussian $\mathcal{H}_i$ consists of the Gaussian parameters used in Gaussian splatting~\cite{kerbl20233d}, along with a link probability $\bm \rho_i$ that represents the likelihood of belonging to each link.
    (\textit{Screw joints}) Each screw joint $\mathcal{S}_j$ consists of a screw axis $\bm S_j$ representing a revolute joint together with the joint spring stiffness $\kappa_j$ and damping coefficient $d_j$. The screw joints are serially connected to form an open chain.
    (\textit{Link inertial parameters}) The parameters $\mathcal{M}_j$ consist of the link mass $m_j$ and rotational inertia matrix $\bm I_j$, and the friction coefficient $\mu_j$ between the link and the supporting surface.
    (\textit{Joint configuration}) The state of the open-chain system at timestep $k$ is given by its joint configuration $\bm \theta_k$.    
    }
    \vspace{-10pt}
    \label{fig:core_components}
\end{figure*}

In this work, we focus on manipulation scenarios in which a robot continuously grasps one end of a DLO using a single arm throughout the interaction.
Consequently, the base of the articulated structure is neither fixed nor floating, but follows a trajectory $\bm T_b(t)\in\SEthree$ prescribed by the robot end-effector.
We assume access to synchronized multi-view RGB observations of the robot--DLO interaction from calibrated cameras and to the corresponding robot end-effector poses.

To avoid ambiguity in notation, we establish the following indexing conventions throughout the paper: The index $i$ denotes Gaussians, $j$ denotes screw axes, and $k$ denotes the discrete timestep index of the multi-view RGB video observations corresponding to a discretization of continuous time $t \in [0, T]$ with video frequency $f$ such that $k = 1, \ldots, \lfloor Tf \rfloor$. 


\subsection{Components of \chainsplat}
\label{subsec:ChainSplat_components}
\chainsplat models a DLO as an open-chain structure, in which a sequence of rigid links is connected by revolute joints to represent its kinematics and dynamics. Each link is associated with a set of 3D Gaussians to capture its geometry and appearance.
Based on this idea, \chainsplat is designed around the four core components illustrated in Fig~\ref{fig:core_components}: screw joints, link inertial parameters, link-aware Gaussians, and joint configurations. First, we define a set of $n_s$ serially-connected non-actuated {\it screw joints}, defining the kinematics of the DLO. The $j$-th screw joint $\mathcal{S}_j$ is parameterized by the tuple of physical parameters
\begin{equation}
\mathcal{S}_j = (\bm S_j, \kappa_j, d_j), 
\end{equation} 
where $\bm S_j \in \mathbb{R}^6$ is the screw axis associated with a revolute joint, while $\kappa_j \in \mathbb{R}_{\geq 0}$ and $d_j \in \mathbb{R}_{\geq 0}$ represent the spring stiffness and damping coefficients of the joint, respectively. Next, we endow each link with a set of {\it inertial parameters} that parametrize the dynamics of the DLO, represented by the tuple
\begin{equation}
\mathcal{M}_j = (m_j, \bm I_j, \mu_j),
\end{equation}
where $m_j \in \mathbb{R}_{>0}$ and $\bm I_j \in \mathbb{R}^{3 \times 3}$ are the mass and rotational inertia matrix of the link, respectively, and $\mu_j \in \mathbb{R}_{\geq 0}$ is the friction coefficient between the link and the supporting surface. The friction coefficients are non-zero and optimized when the DLO is in contact with the environment, e.g., when manipulating DLOs on a tabletop. Together, the screw-joint physical parameters and the link inertial parameters are sufficient to characterize the underlying physical properties of DLOs in most practical scenarios.

Additionally, we define a set of $n_g$ {\it link-aware Gaussians} to represent the link-wise geometry and appearance of the DLO. Compared with the Gaussian primitives introduced in Section~\ref{sec:3dgs}, each link-aware Gaussian additionally includes a link index that specifies the rigid link to which the Gaussian is attached. Accordingly, the $i$-th link-aware Gaussian, denoted by $\mathcal{H}_i$, is parameterized by the augmented tuple
\begin{equation}
\mathcal{H}_i =
(\bm T_i, \bm s_i, \sigma_i, \bm c_i, \bm \rho_i),
\end{equation}
where the parameters $\bm T_i$, $\bm s_i$, $\sigma_i$, $\bm c_i$ are as in~\eqref{eq:gaussian_splatting_components}, and ${\bm \rho_i = (\rho_{i0}, \ldots, \rho_{in_s}) \in \Delta^{n_s}}$ is the probability that the Gaussian $\mathcal{H}_i$ is associated to the $j$-th link (including the base link $0$), with $\Delta^{n_s}$ denoting the $n_s$-dimensional probability simplex. Finally, for each timestep $k$, we define the {\it joint configuration}
\begin{equation}
\bm \theta_k =
(\theta_{k1}, \ldots, \theta_{kn_s})
\in \mathbb{R}^{n_s},
\end{equation}
representing the state of the deformable linear object associated with the RGB observations at timestep $k$. 


\subsection{Differentiable RGB Rendering with \chainsplat} 
\label{subsec:ChainSplat_rgb_render}
\begin{figure}[!t]
    \centering
    \includegraphics[width=\linewidth]{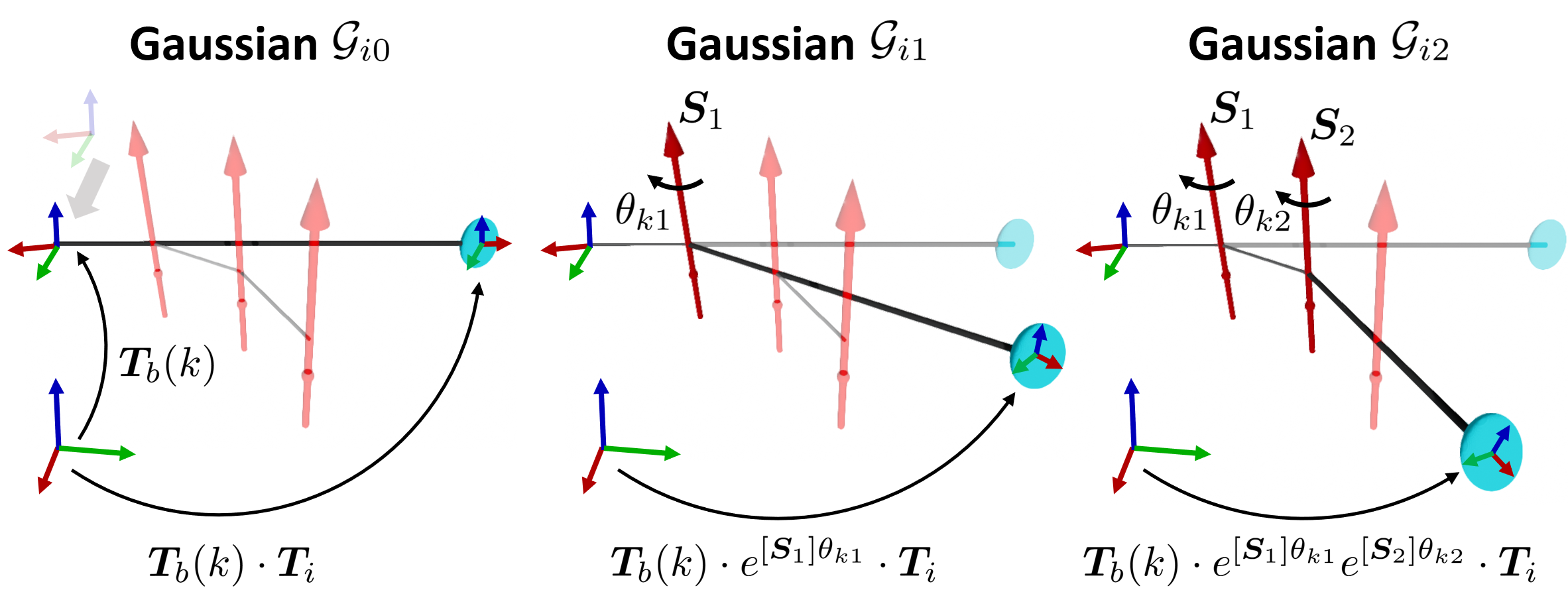}
    \vspace{-15pt}
    \caption{
    Replicated Gaussian primitives derived from $\mathcal{H}_i$. For the base link, the blue Gaussian ellipsoid is transformed as $\bm T_b(k) \bm T_i$. For the first link, the Gaussian is additionally transformed by the first screw motion. For each subsequent link, the Gaussian is transformed according to the product-of-exponentials (PoE) formula.
    }
    \vspace{-15pt}
    \label{fig:chainsplat_render}
\end{figure}

One straightforward approach for rendering an RGB image with \chainsplat given a joint configuration is to assign each link-aware Gaussian $\mathcal{H}_i$ to the link with the highest probability in $\bm \rho_i$, i.e., to the $j_i$-th link with $j_i=\operatorname{argmax}_j \rho_{ij}$, then transform the Gaussian according to the corresponding link motion with~\eqref{eq:poe_formula}, and render the image using the transformed Gaussians. 
However, this hard assignment is not differentiable with respect to the link assignment probabilities $\bm \rho_i$, resulting in a non-smooth optimization landscape that can lead to poor local minima.
To enable differentiable rendering while allowing smooth optimization of $\bm \rho_i$, we adopt an idea similar to ScrewSplat~\cite{kim2025screwsplat}. 
The key idea is to replicate each link-aware Gaussian $\mathcal{H}_i$ into $(n_s+1)$ Gaussian primitives $\mathcal{G}_{ij}$, $j=\{0,\ldots,n_s\}$, each associated with a link. 
Each replicated Gaussian $\mathcal{G}_{ij}$ follows the transformation of its associated link $j$, while its opacity is scaled by the corresponding link-assignment probability $\rho_{ij}$. Figure~\ref{fig:chainsplat_render} illustrates examples of replicated Gaussians from a link-aware Gaussian.
Given $n_g$ link-aware Gaussians, this procedure yields $n_g(n_s+1)$ replicated Gaussian primitives, which are used to render an RGB image through differentiable $\alpha$-blending as in~\eqref{eq:alpha_blending}. This formulation allows the optimization variables including link-assignment probabilities of each Gaussian to be smoothly optimized from ground-truth RGB videos.

When computing the link transformations, we also account for the base-link transformation, as the base moves according to the robot end-effector motion.
Let $\bm T_b(k)\in\SEthree$ denote the base pose at timestep $k$.
Specifically, for the base link ($j=0$), the replicated Gaussian is defined as
\begin{equation}
\mathcal{G}_{i0}
=
(
\bm T_b(k) \cdot \bm T_i,\,
\bm s_i,\,
\sigma_i \rho_{i0},\,
\bm c_i
).
\label{eq:base_link_replicate}
\end{equation}
For the other links ($j \geq 1$), the Gaussian pose is transformed according to the product-of-exponentials forward kinematics in~\eqref{eq:poe_formula}. Given a joint configuration $\bm \theta_k$, the replicated Gaussian is parameterized as
\begin{equation}
\mathcal{G}_{ij}
=
\left(
\bm T_b(k)
\cdot
\prod_{l=1}^{j}
e^{[\bm S_l]\theta_{kl}}
\cdot
\bm T_i,\,
\bm s_i,\,
\sigma_i \rho_{ij},\,
\bm c_i
\right).
\label{eq:link_replicate}
\end{equation}
The scale $\bm s_i$ and appearance $\bm c_i$ of each replicated Gaussian are inherited from the original link-aware Gaussian $\mathcal{H}_i$. The opacity of each replicated Gaussian is modulated by the corresponding part assignment probability $\rho_{ij}$.

This differentiable rendering formulation enables joint optimization of the geometry, appearance, and joint configurations from multi-view RGB videos. 
Moreover, given a camera pose, this formulation defines a differentiable rendering function $\pi$ that maps a joint configuration $\bm\theta$ to an RGB image $\mathcal{I}$, i.e., $\mathcal{I}=\pi(\bm\theta)$.
This property is particularly useful for DLO state estimation, which will be discussed in Section~\ref{subsec:state_estimation}.


\subsection{Dynamics of \chainsplat}
\label{subsec:ChainSplat_dynamics}
Next, we construct the DLO dynamics model based on the core components of \chainsplat and on the equations of motions~\eqref{eq:lie_group_dynamics} for open-chain system introduced in Section~\ref{sec:dynamics_open_chain}. 

To do so, we introduce several additional modeling components.
First, since we assume that the robot continuously grasps one end of the DLO, the base link follows the trajectory prescribed by the robot end-effector.
Let $\bm T_b(t)\in\SEthree$ denote the trajectory of the grasped end of the DLO. The spatial velocity $\bm V_b(t)\in\sethree$ and acceleration $\dot{\bm V}_b(t)\in\sethree$ are then obtained from~\eqref{eq:spatial_velocity}.
To account for the moving base in the dynamics equation, it suffices to specify the base spatial velocity $\bm V_0$ and acceleration $\dot{\bm V}_0$ in the forward iteration~\eqref{eq:newton_euler_forward_1}-\eqref{eq:newton_euler_forward_2}, which we set as $\bm V_0=\bm V_b$ and $\dot{\bm V}_0=(\bm 0,-\bm g)+\dot{\bm V}_b$.

Next, to capture the passive behavior of the DLO joints, we model each revolute joint as a torsional spring--damper system, $\tau_j=-\kappa_j\theta_j-d_j\dot{\theta}_j$, 
or, equivalently, in matrix form, 
\begin{equation}
\bm\tau=-\bm K\bm\theta-\bm D\dot{\bm\theta}
\end{equation}
where $\bm K=\operatorname{diag}(\kappa_1,\ldots,\kappa_{n_s})$, $\bm D=\operatorname{diag}(d_1,\ldots,d_{n_s})\in\mathbb{R}^{n_s\times n_s}$ denote the joint stiffness and damping matrices, respectively.

Finally, when the DLO is in contact with a supporting surface, such as a table, friction becomes a dominant external forceand should therefore be explicitly incorporated into the dynamics.
To this end, we introduce a link-wise viscous friction model into the dynamics. Specifically, the backward iteration in~\eqref{eq:newton_euler_backward} is extended as
\begin{equation}
\bm F_i
 =
[\Adjoint_{\bm T_{i+1, i}}]^{\trsp}
\bm F_{i+1}
+
\bm G_i \dot{\bm V}_i
-
[\adjoint_{\bm V_i}]^{\trsp}\bm G_i \bm V_i
+
\bm \mu_i \bm V_i,
\label{eq:viscous_friction_model}
\end{equation}
where the additional term $\bm\mu_i\bm V_i$ models translational viscous friction, with $\bm\mu_i=\operatorname{diag}(0,0,0,\mu_i,\mu_i,\mu_i)\in\mathbb{R}^{6\times6}$.

Following the same derivation used to obtain the closed-form dynamics in~\eqref{eq:inverse_dynamics_mass}-\eqref{eq:inverse_dynamics_gravity}, the dynamics of \chainsplat are expressed in closed-form as
\begin{equation}
\begin{aligned}
-\bm K \bm \theta -\bm D \dot{\bm \theta}
&=
\bm M(\bm \theta)\ddot{\bm \theta}
+
\bm C(\bm \theta, \dot{\bm \theta})\dot{\bm \theta}
+
\bm g(\bm \theta)
\\
&\quad
-
\bm M_0(\bm \theta)\dot{\bar{\bm V}}_b
-
\bm C_0(\bm \theta, \dot{\bm \theta})\bar{\bm V}_b
+
\bm \mu \bm V,
\label{eq:chainsplat_dynamics}
\end{aligned}
\end{equation}
where $\bm M_0(\bm\theta)$ and $\bm C_0(\bm\theta,\dot{\bm\theta})$ capture the dynamic coupling induced by the robot-controlled base motion and are given by
\begin{align} 
\bm M_0(\bm \theta) 
& = 
\bm S^\top \bm L(\bm \theta)^\top \bm G \bm L(\bm \theta), 
\\ 
\bm C_0(\bm \theta, \dot{\bm \theta}) 
& = \bm S^\top \bm L(\bm \theta)^\top \left( \bm G \bm L(\bm \theta) [\adjoint_{\bm S \dot{\bm \theta}}] + [\adjoint_{\bm V}]^{\trsp} \bm G \right) \bm L(\bm \theta), \end{align}
respectively, and the vector $\bar{\bm V}_b$ is defined in~\eqref{eq:bar_operation}. The term $\bm\mu\bm V$ represents the generalized viscous friction friction acting on the joints, where $\bm V$ is the stacked spatial velocities of all links described in~\eqref{eq:stacked_velocity}. The matrix 
\begin{equation}
    \bm \mu = \bm S^\top \bm L(\bm \theta)^\top \mathrm{diag}(\bm \mu_1, \ldots, \bm \mu_{n_s})
\end{equation}
maps the link-wise viscous frictional forces to joint configurations.

The \chainsplat dynamics formulation offers two key advantages: (1) it enforces physical consistency through an explicit physics model, and (2) it yields an efficient low-dimensional dynamics model, in which the system state is parameterized solely by the joint configurations of the open-chain structure.

%% file: 05optimization/v1.tex

\section{Optimizing the Parameters of ChainSplat from Multi-view RGB Videos}
\label{sec:chainsplat_optimization}
In this section, we present our framework for estimating the parameters of \chainsplat from synchronized multi-view RGB videos. We formulate the learning problem as a two-stage optimization procedure consisting of a \textit{kinematics-aware geometry recognition} stage and a \textit{dynamics identification} stage.
For the first stage, presented in Section~\ref{subsec:kinematics-geometry-recognition}, we design a differentiable RGB rendering formulation that enables fully differentiable end-to-end optimization of the geometry, appearance, and joint configurations of a \chainsplat model by minimizing the discrepancy between the rendered and ground-truth RGB observations. As a byproduct of this stage, we also recover the joint configuration trajectories of the open-chain structure.
In the second stage, presented in Section~\ref{subsec:dynamic_identification}, we estimate the dynamic parameters of \chainsplat by minimizing the discrepancy between the simulated joint configuration trajectories and those recovered in the first stage. The complete optimization procedure is summarized in Algorithm~\ref{algo:chainsplat_optimization}.

\begin{algorithm}[t]
\small
\caption{\chainsplat Optimization and Differentiable Rendering}
\label{algo:chainsplat_optimization}

\SetKwInput{KwInit}{Initialize}
\SetKwFunction{Render}{render}
\SetKwFunction{Recognition}{kin}
\SetKwFunction{Identification}{dyn}
\SetKwProg{Fn}{Function}{:}{}

\KwIn{Multi-view RGB videos $\{\mathcal{I}_k\}_{k=1}^{T}$,
base trajectory $\bm T_b$,
predefined screw axes $\{\bm S_j\}_{j=1}^{n_s}$
}

\KwOut{Link-aware Gaussians $\{\mathcal{H}_i\}_{i=1}^{n_g}$,
joint configurations $\{\bm{\theta}_k\}_{k=1}^{T}$,
dynamic parameters $\{m_j,d_j,\mu_j\}_{j=1}^{n_s}$
}

\textcolor{softil}{$\{\mathcal{H}_i\}_{i=1}^{n_g},
\{\bm{\theta}_k\}_{k=1}^{T}
\leftarrow$
\Recognition{$\{\mathcal{I}_k\}_{k=1}^{T}$,
    $\bm T_b$,
    $\{\bm S_j\}_{j=1}^{n_s}$}
}

\textcolor{softst}{$\{m_j,d_j,\mu_j\}_{j=1}^{n_s}
\leftarrow$
\Identification{$\{\bm{\theta}_k\}_{k=1}^{T}$,
    $\bm T_b$}
}

\algosection[softil]{Kinematics-aware Geometry Recognition}

\Fn{\Recognition{$\{\mathcal{I}_k\}_{k=1}^{T}$,
    $\bm T_b$,
    $\{\bm S_j\}_{j=1}^{n_s}$
}}{

\KwInit{Link-aware Gaussians $\{\mathcal{H}_i\}_{i=1}^{n_g}$
and joint configurations
$\{\bm{\theta}_k\}_{k=1}^{T}$
}

\While{not converged}{

    Randomly select camera $l \in \{1, 2, 3\}$

    Randomly sample $k \in \{1,\ldots,T\}$.

    \textcolor{softpi}{
    $\hat{\mathcal{I}}_k^l
    \leftarrow$
    \Render{$\{\mathcal{H}_i\}_{i=1}^{n_g}$,
        $\{\bm S_j\}_{j=1}^{n_s}$,
        $\bm{\theta}_k$,
        $\bm T_b(k)$
    }
    }

    Compute
    $\mathcal{L}_{\mathrm{render}}
    (\hat{\mathcal{I}}_k^l,\mathcal{I}_k^l)$.
    \hfill Eq.~\eqref{eq:rendering_loss}

    Update
    $\{\mathcal{H}_i\}_{i=1}^{n_g}$
    and $\bm{\theta}_k$
    by minimizing
    $\mathcal{L}_{\mathrm{render}}$.
}

\Return{$\{\mathcal{H}_i\}_{i=1}^{n_g},
\{\bm{\theta}_k\}_{k=1}^{T}$
}
}

\algosection[softst]{Dynamics Identification}

\Fn{\Identification{$\{\bm{\theta}_k\}_{k=1}^{T}$,
    $\bm T_b$
}}{

\KwInit{Dynamic parameters
$\{m_j,d_j,\mu_j\}_{j=1}^{n_s}$
}

\While{not converged}{

    Set initial $\hat{\bm{\theta}}_1 = \bm{\theta}_1$.

    Get $\{\hat{\bm{\theta}}_k\}_{k=1}^{T}$ integrating \chainsplat dynamics under
    base trajectory $\bm T_b$. 
    \hfill Eq.~\eqref{eq:chainsplat_dynamics}

    Compute
    $\mathcal{L}_{\mathrm{dynamics}}
    \left(
        \{\hat{\bm{\theta}}_k\}_{k=1}^{T},
        \{\bm{\theta}_k\}_{k=1}^{T}
    \right)$.
    \hfill Eq.~\eqref{eq:dynamics_loss}

    Update
    $\{m_j,d_j,\mu_j\}_{j=1}^{n_s}$
    by minimizing
    $\mathcal{L}_{\mathrm{dynamics}}$.
}

\Return{$\{m_j,d_j,\mu_j\}_{j=1}^{n_s}$
}
}

\algosection[softpi]{Differentiable \chainsplat Rendering}

\Fn{\Render{$\{\mathcal{H}_i\}_{i=1}^{n_g}$,
    $\{\bm S_j\}_{j=1}^{n_s}$,
    $\bm{\theta}$,
    $\bm T_0$
}}{

\For{$i=1,\ldots,n_g$}{

    Construct the Gaussian $\mathcal{G}_{i0}$
    from $\mathcal{H}_i$.
    \hfill Eq.~\eqref{eq:base_link_replicate}

    Construct the Gaussians $\mathcal{G}_{ij}$
    from $\mathcal{H}_i$ for $j=1,\ldots,n_s$.
    \hfill Eq.~\eqref{eq:link_replicate}

}

Collect all replicated Gaussians
$\{\mathcal{G}_{ij}\}_{i=1,j=0}^{n_g,n_s}$.

Render $\hat{\mathcal{I}}$ using
differentiable $\alpha$-blending.
\hfill Eq.~\eqref{eq:alpha_blending}

\Return $\hat{\mathcal{I}}$
}

\end{algorithm}


\subsection{Kinematics-aware Geometry Recognition}
\label{subsec:kinematics-geometry-recognition}
The objective of this stage is to optimize the geometry, appearance, and kinematics of \chainsplat from multi-view RGB video observations. 
Using the differentiable RGB rendering framework described in Section~\ref{subsec:ChainSplat_rgb_render}, we optimize the components of \chainsplat by minimizing the discrepancy between the rendered and ground-truth RGB videos. As ground-truth observations, we use object-masked RGB videos, where the object masks are obtained using the Segment Anything Model (SAM)~\cite{kirillov2023segment}.

\subsubsection{Optimization Variables for Kinematics-aware Geometry Recognition}
The components involved in the kinematics-aware geometry are the screw axes $\bm{S}_j$ of the screw joints $\mathcal{S}_j$, the link-aware Gaussians $\mathcal{H}_i$, and the joint configurations $\bm{\theta}_k$.
Although all these parameters could, in principle, be optimized jointly, we introduce a structural constraint to improve optimization stability. 
Specifically, we empirically found that optimizing the screw axes $\bm S_j$ jointly with the remaining parameters often leads to physically implausible articulated structures despite achieving low RGB reconstruction errors, resulting in poor generalization and unstable dynamics estimation. Therefore, we constrain all screw axes $\bm S_j$ to point to the same direction and place them at uniformly spaced locations along the DLO. The screw axes are then kept fixed throughout the optimization. This assumption is appropriate for a broad class of DLO manipulation tasks in which deformations are mostly limited to a plane, e.g., DLO manipulation on a tabletop or in free space with linear base motions.
Therefore, the remaining optimization variables for this stage are link-aware Gaussians $\mathcal{H}_i$ and the joint configurations $\bm{\theta}_k$.

\subsubsection{Optimization Details} 
The first four parameters link-aware Gaussians $\mathcal{H}_i=(\bm T_i,\bm s_i,\sigma_i,\bm c_i,\bm\rho_i)$ are initialized following~\cite{kerbl20233d} with uniformly sampled poses $\bm T_i$, small isotropic scales $\bm s_i$, opacities $\sigma_i$ of $0.1$, and random colors $\bm c_i$, while the link-assignment probabilities $\bm\rho_i\in\Delta^{n_s}$ are initialized to a uniform distribution. 
All joint configurations $\bm \theta_k$ are initialized to zero vectors.

For the training dataset, we use a single set of multi-view RGB videos capturing robot--object interactions under predesigned robot end-effector motions. 
The link-aware Gaussians $\mathcal{H}_i$ and joint configurations $\bm {\theta}_k$ are then optimized by minimizing the rendering objective $\mathcal{L}_{\mathrm{render}}$~\eqref{eq:rendering_loss} between the RGB images rendered by \chainsplat and the corresponding observations.

During optimization, we periodically reset the Gaussian opacities to values close to zero, following~\cite{kerbl20233d}, to encourage the discovery of more informative Gaussian primitives and mitigate poor local minima. 
In addition, we further introduce a periodic link probability reset in \chainsplat, in which the link probabilities $\bm{\rho}_i$ are reinitialized to uniform distributions throughout training. Importantly, the link probability reset is performed asynchronously with respect to the opacity reset schedule used in Gaussian Splatting. Empirically, we find that combining these two reset strategies significantly improves optimization stability and leads to a more reliable decomposition of articulated links.

After optimization, we collect $\bm\theta_k$ across all timesteps $k$ to obtain the joint configuration trajectory of the DLO from RGB videos. This trajectory is then used to identify the dynamics parameters of \chainsplat in the subsequent section.


\subsection{Dynamics Identification}
\label{subsec:dynamic_identification}
The objective of this second stage is to estimate the dynamic parameters of \chainsplat. Given the robot-controlled base trajectory, we simulate the open-chain dynamics by integrating~\eqref{eq:chainsplat_dynamics}. 
The dynamic parameters are then optimized by minimizing the discrepancy between the simulated joint trajectories and those obtained from the preceding kinematics-aware geometry recognition stage.

\subsubsection{Optimization Variables for Dynamics Identification}
The parameters involved in the dynamics model are the spring--damper coefficients $k_j$ and $d_j$ associated with the screw joints $\mathcal{S}_j$, and the link inertial parameters $\mathcal{M}_j = (m_j, \bm I_j, \mu_j)$. 
We further introduce several constraints to improve optimization stability. 
We assume that each link is a uniform-density rod, which is reasonable for typical DLOs. Accordingly, the rotational inertia about the screw axis is given by $\frac{1}{12 }m_j l^2$, where $l$ is the distance between adjacent screw joints.
We focus on DLOs with negligible elasticity, e.g., cables and ropes, and set the spring stiffness of all screw joints to zero, i.e., $k_j=0$.
Consequently, the optimization variables for dynamics identification are the damping coefficients $d_j$, link masses $m_j$, and, for tabletop scenarios only, friction coefficients $\mu_j$.

\subsubsection{Dynamics Simulation} Given the joint configuration $\bm \theta_k$ and velocity $\dot{\bm \theta}_k$, the articulated dynamics are integrated using the explicit Euler method as
\begin{equation}
\label{eq:explicit_euler}
\dot{\bm \theta}_{k+1}
=
\dot{\bm \theta}_{k}
+
h \ddot{\bm \theta}_{k},
\:\:\:
\bm \theta_{k+1}
=
\bm \theta_{k}
+
h \dot{\bm \theta}_{k},
\end{equation}
where $h$ is the simulation timestep. However, when the timestep is not sufficiently small --- e.g., for a camera framerate of $30$ Hz --- the explicit integration~\eqref{eq:explicit_euler} often suffers from numerical instability, particularly in dynamical systems containing velocity-dependent forces such as damping terms. One possible, although computationally-expensive, solution is to decrease $h$ by downsampling the trajectories.

Instead, we adopt the implicit Euler method, whose integration step is
\begin{equation}
\dot{\bm \theta}_{k+1}
=
\dot{\bm \theta}_{k}
+
h \ddot{\bm \theta}_{k+1},
\:\:\:
\bm \theta_{k+1}
=
\bm \theta_{k}
+
h \dot{\bm \theta}_{k+1}.
\end{equation}
Since the acceleration at the next timestep depends implicitly on the unknown future state, directly evaluating $\ddot{\bm \theta}_{k+1}$ requires solving a nonlinear system. To obtain an efficient approximation, we employ the {\it implicit-in-velocity Euler} method~\cite{wanner1996solving}, which uses a first-order Taylor expansion with respect to the velocity term $\dot{\bm \theta}_k$.
Specifically, by representing the acceleration as a function of the velocity, i.e., $\ddot{\bm \theta}_k = a(\dot{\bm \theta}_k)$, we obtain
\begin{equation}
\begin{aligned}
\dot{\bm \theta}_{k+1}
&=
\dot{\bm \theta}_{k}
+
h a(\dot{\bm \theta}_{k+1})
\\
&\approx
\dot{\bm \theta}_{k}
+
h
\left(
a(\dot{\bm \theta}_{k})
+
\frac{\partial a(\dot{\bm \theta}_{k})}
{\partial \dot{\bm \theta}_{k}}
(\dot{\bm \theta}_{k+1} - \dot{\bm \theta}_{k})
\right).
\end{aligned}
\end{equation}
Rearranging the above expression yields the implicit-in-velocity Euler update
\begin{equation}
\begin{aligned}
\dot{\bm \theta}_{k+1}
&=
\dot{\bm \theta}_{k}
+
h
\left(
\bm I
-
h
\frac{\partial a(\dot{\bm \theta}_{k})}
{\partial \dot{\bm \theta}_{k}}
\right)^{-1}
\ddot{\bm \theta}_{k},
\\
\bm \theta_{k+1}
&=
\bm \theta_k
+
h \dot{\bm \theta}_{k+1}.
\end{aligned}
\end{equation}

An important advantage of \chainsplat is that the Jacobian
$\partial a(\dot{\bm \theta}) / \partial \dot{\bm \theta}$
admits a closed-form expression, thereby enabling efficient implicit-in-velocity Euler integration without requiring numerical differentiation during simulation. Leveraging \chainsplat's formulation, the Jacobian is given by
\begin{equation}
\label{eq:closed-form-jacobian}
\frac{\partial a(\dot{\bm \theta})}
{\partial \dot{\bm \theta}}
=
\bm M(\bm \theta)^{-1}
\hat{\bm M},
\end{equation}
with
\begin{equation}
\begin{aligned}
\hat{\bm M}
=&
-\bm D
+
\bm \mu \bm L(\bm \theta)\bm S
+
\bm S^{\trsp}
\bm L(\bm \theta)^{\trsp}
\Big(
-\bm Q \bm L(\bm \theta)\bm S
\\
&+
\bm G \bm L(\bm \theta)
[\adjoint_{\bm V}]
\bm S
-
[\adjoint_{\bm G\bm V}^{\star}]
\bm L(\bm \theta)\bm S
\Big),
\\
\bm Q
=&
\bm G \bm L(\bm \theta)
[\adjoint_{\bm S\dot{\bm \theta}}]
+
[\adjoint_{\bm V}]^\trsp
\bm G,
\end{aligned}
\end{equation}
where $\adjoint_{\bm F}^{\star} : \sethree^* \rightarrow \sethree^*$
denotes the dual adjoint (or coadjoint) operator associated with the spatial
wrench $\bm F = (\bm m,\bm f)$ defined by
\begin{equation}
[\adjoint_{\bm F}^{\star}]
=
\left(
\begin{matrix}
[\bm m] & [\bm f] \\
[\bm f] & \bm 0
\end{matrix}
\right).
\end{equation}

\subsubsection{Optimization Details} 
The damping coefficients $d_j$, link masses $m_j$, and, for tabletop scenarios, friction coefficients $\mu_j$, are initialized to the same values across all links $j$. These parameters are optimized by minimizing the discrepancy between the simulated joint trajectories and those obtained from the kinematics-aware recognition stage. Specifically, we minimize the sum-of-squares objective
\begin{equation}
\mathcal{L}_{\mathrm{dynamics}}
=
\sum_k
\left\|
\hat{\bm \theta}_k
-
\bm \theta_k
\right\|^2,
\label{eq:dynamics_loss}
\end{equation}
where $\hat{\bm \theta}_k$ and $\bm \theta_k$ denote the simulated and ground-truth joint configurations at timestep $k$, respectively.

%% file: 06manipulation/v3.tex
\section{Model-based Manipulation with \chainsplat}
\label{sec:chainsplat_manipulation}
In this section, we build on the optimized \chainsplat representation to develop a model-based manipulation framework for DLOs. To do so, we leverage two key advantages of \chainsplat: (1) its low-dimensional dynamics representation, where the system state is parameterized by the joint configurations of the open-chain structure, and (2) its differential rendering ability, characterized by the rendering function $\pi$ that maps a joint configuration $\bm \theta$ to an RGB image $\mathcal{I}$, see Section~\ref{subsec:ChainSplat_rgb_render}.
The former enable efficient model-based trajectory optimization for manipulation tasks (Section~\ref{subsec:trajectory_optimization}), while the latter allows the estimation of the DLO state directly from RGB observations (Section~\ref{subsec:state_estimation}). Moreover, by minimizing discrepancies between predicted and observed DLO states, the framework can estimate external contact positions and interaction forces applied to the object (Section~\ref{subsec:force_estimation}).

\begin{figure}[!t]
    \centering
    \includegraphics[width=\linewidth]{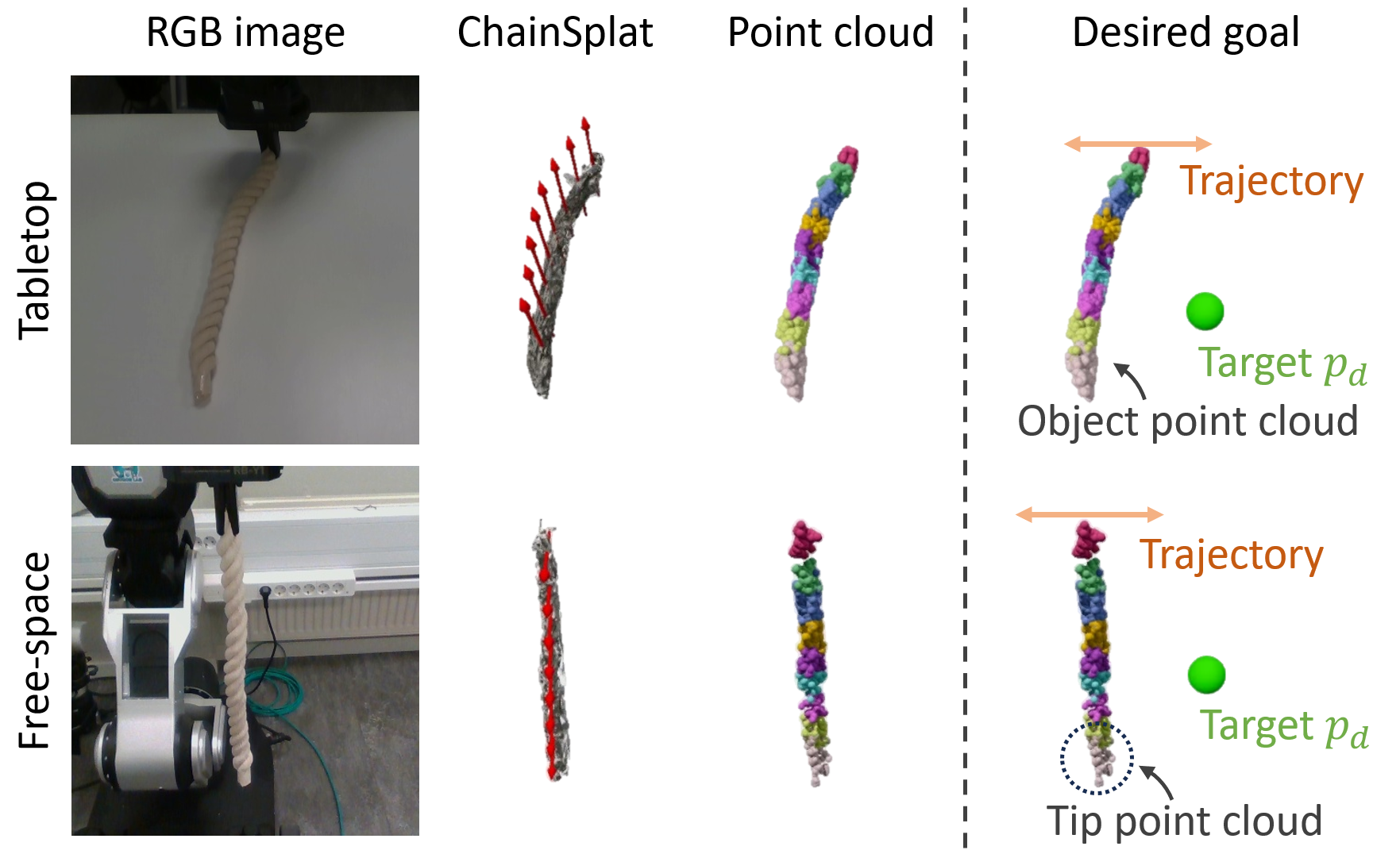}
    \vspace{-15pt}
    \caption{
    Illustration of the objective function for the hitting-target manipulation task. 
    (\textit{Left}) A differentiable point-cloud function with part assignments represents the 3D shape of the DLO obtained from \chainsplat. 
    (\textit{Right}) The objective is to hit the target point with any point of the DLO point cloud for tabletop scenarios and with the tip point cloud (i.e., points belonging to the last link) for free-space scenarios.
    }
    \vspace{-15pt}
    \label{fig:model_based_manipulation}
\end{figure}

\subsection{Model-based Trajectory Optimization}
\label{subsec:trajectory_optimization}
Trajectory optimization within \chainsplat aims to find an optimal robot-controlled trajectory $\bm T_b^\star(t)$ for the base of the DLO --- i.e., the end of the DLO grasped by the robot --- so that it minimizes a task-specific control objective. In this work, we consider a simple, yet highly dynamic manipulation task: Hitting a fixed desired target point $\bm p_d \in \mathbb{R}^3$ in 3D space. We consider two variants of this task, shown in Figure~\ref{fig:model_based_manipulation}: (1) a tabletop setting, where frictional interactions dominate the dynamics, and (2) a free-space setting, where the dynamics are primarily governed by gravity.

For trajectory optimization, two components must be specified, namely an appropriate parameterization of the robot end-effector trajectory and an objective function corresponding to the desired manipulation task. We restrict the end-effector motion to a one-dimensional trajectory, consistent with the base motions used during training. This restriction also avoids trivial solutions in which the end-effector simply moves the DLO to the target location, encouraging the optimizer to exploit the dynamic behavior of the DLO. Following~\cite{lee2024differentiable}, we parameterize the trajectory using radial basis functions (RBFs). Specifically, we parameterize the end-effector trajectory as
\begin{align}
&\bm T_b(t; \bm{w}, x_T) =\left(\bm I_3, (x(t),0,0)^\trsp\right) \quad \text{with} \\
&x(t)
=
x_0
+
(x_T - x_0)(3 - 2s)s^2
+
s^2(s-1)^2 \Phi(s)\bm{w}, \nonumber
\end{align}
where $s = \frac{t}{T}$, and $x_0, x_T \in \mathbb{R}$ are the initial and terminal positions, respectively, $\Phi$ is a basis function matrix defined as
$\Phi(s)
=
\begin{bmatrix}
\phi_1(s) & \cdots & \phi_B(s)
\end{bmatrix}
\in \mathbb{R}^{1 \times B},
$
with Gaussian basis functions
\begin{equation}
    \phi_i(s)
=
\exp\!\left(
-B^2
\left(
s - \frac{i-1}{B-1}
\right)^2
\right),
\end{equation}
for $i = 1, \ldots, B$, and $\bm{w}\in\mathbb{R}^{B}$ is the coefficient vector. Note that other trajectory parameterizations (e.g., spline-based representations) are equally applicable within our framework.  

Given an objective function $c(\bm T_b)$ defined over the end-effector trajectory, the trajectory optimization problem is formulated as
\begin{equation}
    \bm T_b^*(t; \bm{w}^*, x_T^*) = \operatorname*{argmin}_{\bm{w}, x_T} c\big(\bm T_b(t; \bm{w}, x_T)\big).
\end{equation}
Here, we formulate the objective function $c$ directly in the 3D workspace based on the reconstructed geometry of \chainsplat. 
Specifically, we first extract the points $\bm p_i$ from the poses of the link-aware Gaussians $\mathcal{H}_i$, which collectively represent the 3D geometry of the DLO, and assign each Gaussian to the link with the highest assignment probability, i.e., $j_i=\operatorname{argmax}_j \rho_{ij}$.
Given a base trajectory $\bm T_b(t)$, we integrate the \chainsplat dynamics~\eqref{eq:chainsplat_dynamics} to obtain the corresponding joint configuration trajectory $\bm \theta_k$'s. 
At each timestep $k$, the collected points are then transformed via forward kinematics according to their assigned links $j_i$, yielding the 3D position $\bm p_i(k;\bm T_b)$ of each point at timestep $k$. 
Importantly, $\bm p_i(k;\bm T_b)$ is differentiable with respect to the base trajectory $\bm T_b$, enabling gradient-based trajectory optimization.

For the tabletop setting, we design a trajectory optimization objective that encourages any point on the DLO to reach a desired target point $\bm p_d \in \mathbb{R}^3$ in the workspace. The objective is formulated as
\begin{equation}
c(\bm T_b)
=
\min_i
\|
\bm p_i(\lfloor Tf \rfloor; \bm T_b) - \bm p_d
\|^2
+
\alpha
\sum_k
\|
\dot{\bm V}_b(k)
\|^2,
\label{eq:tabletop_objective}
\end{equation}
where $T$ is the time horizon, $f$ is the sampling frequency, and $\alpha$ is a weighting coefficient. The first term minimizes the distance between the reconstructed DLO geometry and the target point, while the second term regularizes the smoothness of the generated motion trajectory.
In the free-space setting, stabilization at a fixed target location is generally infeasible due to the highly dynamic nature of the DLO motion. Accordingly, we instead define the objective as
\begin{equation}
c(\bm T_b)
=
\min_k
\min_{i \in \{i | j_i = n_s\}}
\|
\bm p_i(k, \bm T_b) - \bm p_d
\|^2
+
\beta
\sum_k
\|
\dot{\bm V}_b(k)
\|^2,
\label{eq:freespace_objective}
\end{equation}
where $\beta$ is a weighting coefficient. This objective encourages the ungrasped end of the DLO to reach the desired target point $\bm p_d$ at least once along the trajectory.

\subsection{State Estimation}
\label{subsec:state_estimation}
The model-based trajectory optimization described in the previous section is inherently an open-loop control strategy. In practice, however, robust manipulation requires closed-loop control to compensate for modeling errors, external disturbances, and sensing noise. A key prerequisite for closed-loop control is accurate real-time state estimation. Although a complete closed-loop manipulation system is beyond the scope of this work, we propose a state estimation framework for DLOs as an essential building block toward robust closed-loop control, thereby demonstrating the potential of \chainsplat for closed-loop manipulation in future real-world deployments.

Our state estimation framework recovers the DLO joint configuration $\bm \theta$ from RGB observations based on the optimized kinematic structure of \chainsplat. Unlike the training stage of \chainsplat, which requires multi-view RGB videos, state estimation can be performed using only a single RGB image as follows.
Given a new RGB observation $\mathcal{I}_{k+1}$ at timestep $k+1$, we estimate the corresponding joint configuration $\bm \theta_{k+1}$ by minimizing the rendering loss between the observed image and the rendered image generated by \chainsplat, i.e.,
\begin{equation}
\bm \theta_{k+1}^{\star}
=
\operatorname*{argmin}_{\bm \theta}
\mathcal{L}_{\mathrm{render}}
\left(
\pi(\bm \theta),
\mathcal{I}_{k+1}
\right),
\label{eq:state_estimation}
\end{equation}
where $\pi(\bm\theta)$ is the RGB rendering function of \chainsplat.
For real-time state estimation, we set the initial value of the optimization variable $\bm\theta$ as the previously estimated joint configuration $\bm \theta_k$. Since the joint configuration varies smoothly between consecutive video frames, this initialization provides a strong prior allowing the optimization to converge rapidly in practice, thereby enabling near real-time state estimation from streaming RGB observations.

\subsection{External Force Estimation}
\label{subsec:force_estimation}
Finally, \chainsplat's physics-based dynamics model enables the estimation of external forces acting on DLOs, extending the applicability of \chainsplat to contact-aware manipulation scenarios involving interactions with the environment or with humans, without requiring dedicated force/torque sensors.

We aim to estimate the interaction force given an external contact applied to the object.
For this setting, we assume that the external interaction consists only of a translational force, i.e., 
$ \bm F_{\mathrm{ext}} =
(\bm 0, \bm f_{\mathrm{ext}}) \in \mathbb{R}^6$ applied at a single contact point.
Although the proposed formulation naturally extends to manipulation scenarios with a moving robot, in this work we consider the simpler case in which the base of the DLO is fixed, i.e., $\bm V_b=\bm 0$, and where external forces are applied by a human hand.
We further assume that the human hand is sufficiently slow to be considered quasi-static, such that inertial effects are negligible, i.e., $\ddot{\bm\theta}\approx\bm 0$.
Under this assumption, the dynamics, governed by~\eqref{eq:lie_group_dynamics} and \eqref{eq:chainsplat_dynamics}, are given by
\begin{equation}
-\bm D \dot{\bm \theta}
=
\bm C(\bm \theta, \dot{\bm \theta})\dot{\bm \theta}
+
\bm g(\bm \theta)
+
\bm \mu \bm V
+
\bm J(\bm \theta)^\top
\bm F_{\mathrm{ext}},
\label{eq:force_estimation}
\end{equation}
where $\bm J(\bm \theta)$ denotes the contact Jacobian, which depends on both the contact position and the articulated link associated with the contact point.

To estimate the external force, we first obtain the current joint state $\bm\theta_k$ using the real-time state estimation framework of Section~\ref{subsec:state_estimation}. The joint velocity $\dot{\bm\theta}_k$ is then approximated by finite differences. Simultaneously, we obtain the 3D point cloud of the human hand from RGB-D observations using SAM~\cite{kirillov2023segment} and set the contact point as the center of the Gaussian closest to the reconstructed hand point cloud. Finally, we compute the contact Jacobian $\bm J(\bm\theta)$ by differentiating the forward kinematics associated with the selected Gaussian and estimate and estimate the external force as
\begin{equation}
\bm F_{\mathrm{ext}}
=
\left(\bm J(\bm\theta)^\top\right)^\dagger
\left(
-\bm D\dot{\bm\theta}
-\bm C(\bm\theta,\dot{\bm\theta})\dot{\bm\theta}
-\bm g(\bm\theta)
-\bm\mu\bm V
\right),
\label{eq:force_estimation_2}
\end{equation}
where $(\cdot)^\dagger$ denotes the Moore-Penrose pseudoinverse.

%% file: 07experiments/v1.tex
\section{Experiments}
In this section, we empirically demonstrate that \chainsplat (1) outperforms existing state-of-the-art methods in DLO dynamics learning, 3D geometry reconstruction, and RGB rendering fidelity (Section~\ref{subsec:experiments-dynamics-learning}; (2) is effectively applied to DLO state estimation and interaction force estimation from human--object interactions (Section~\eqref{subsec:experiments-estimation}; and (3) demonstrates practical utility for real-world DLO manipulation through model-based trajectory optimization, particularly for hitting desired target points with DLOs (Section~\ref{subsec:experiment-manipulation}). Additional visualizations of all experimental results are provided in the supplementary video.

\subsection{Experimental Settings}
We first describe the real-world experimental setup for training \chainsplat and evaluating it for model-based manipulation. 
We then describe the data collection procedure and experimental details for the optimization of \chainsplat, followed by the state-of-the-art methods for constructing DLO digital twins that we use as baselines.

\subsubsection{Real-World Experimental Setup}
\begin{figure}
    \centering
    \includegraphics[width=\linewidth]{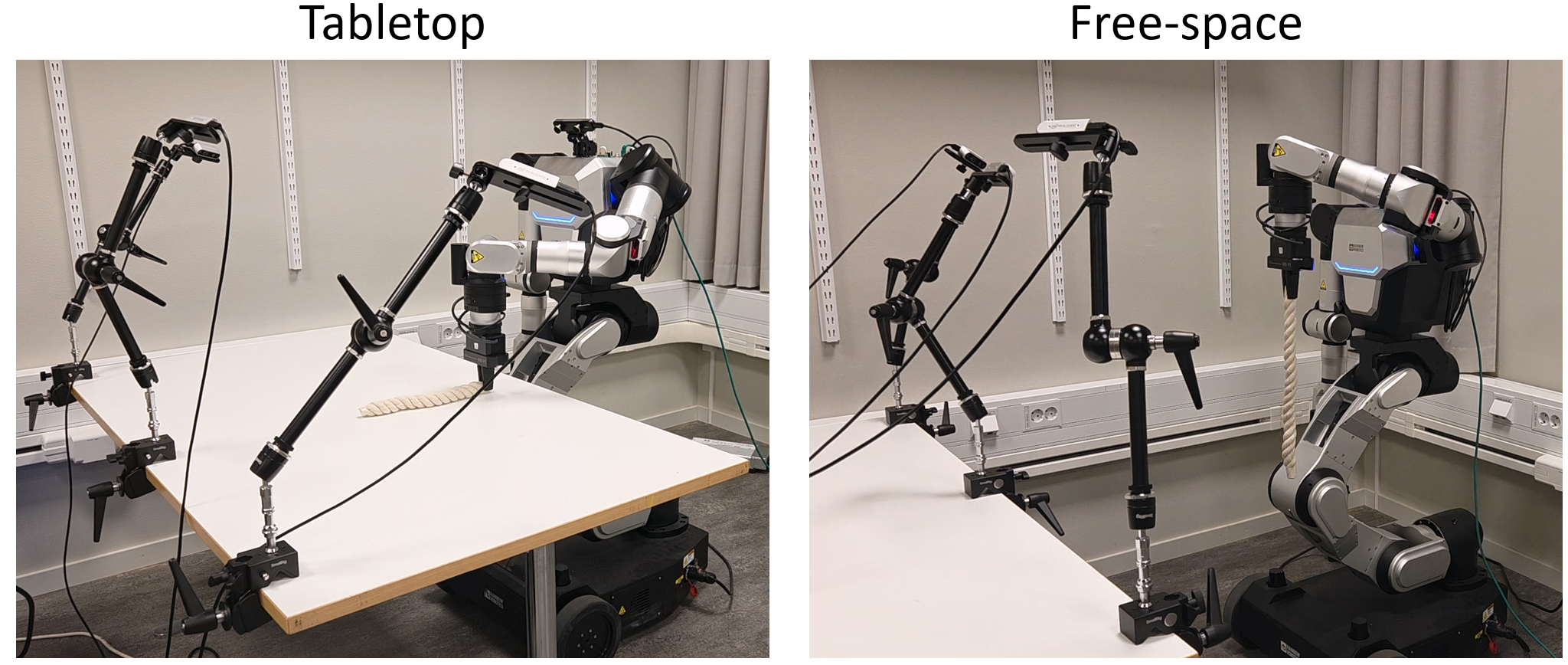}
    \vspace{-15pt}
    \caption{
    Real-world experimental setup for the tabletop and free-space scenarios.
    }
    \vspace{-15pt}
    \label{fig:experimental_setting}
\end{figure}

We use the left arm of the RB-Y1 humanoid robot equipped with a parallel-jaw gripper, along with three Intel RealSense D435 cameras and a ZED Mini RGB-D camera mounted on the robot's head. 
Across both data collection and manipulation experiments, we consider two scenarios, shown in Figure~\ref{fig:experimental_setting}: (1) a tabletop scenario, where frictional interactions dominate the dynamics, and (2) a free-space scenario, where the dynamics are primarily governed by gravity. 
The three RealSense cameras are used for all experiments, while the ZED Mini is used only for calibration to obtain the extrinsic parameters of the RealSense cameras with respect to the robot base frame using AprilTag. 
The base trajectory of the grasped end of the DLO is also expressed in the robot base frame.

\subsubsection{Dataset Collection}
We use a set of three DLOs, namely three $2$cm-thick ropes of lengths $20$~cm, $30$~cm, and $40$~cm. 
For each object, we collect four $5$-second trajectories generated by four distinct robot end-effector motions. 
Specifically, the end-effector follows one-dimensional periodic trajectories along the robot's lateral (left--right) direction, with different motion profiles. 
We use sinusoidal and trapezoidal waveforms with an amplitude of $15$~cm, each executed at a slow ($0.25$~Hz) and fast ($0.4$~Hz) frequency. 
We collect a total of $24$ trajectories across the two experimental scenarios and three DLOs. 
For each trajectory, we synchronously record RGB and depth images at $30$~Hz from the three RealSense cameras, along with the $\SEthree$ base trajectory of the DLO. 
We emphasize that the collected depth images are only used for training the baseline models described hereinafter and for computing evaluation metrics. 

\subsubsection{\chainsplat Optimization}
We use \chainsplat models composed of eight screw joints, corresponding to nine links (including the base link) for all experiments.
For each of the $24$ trajectories, we optimize a separate \chainsplat model and evaluate it on the same base trajectory used for its optimization. 
For kinematics-aware geometry recognition, we use the full $5$-second trajectory recorded at $30$~Hz ($150$ timesteps), subsampling every $5$ timesteps for optimization which yields $30$ joint configuration vectors $\bm\theta_k$ per trajectory. For dynamics identification, we integrate the dynamics at $30$~Hz and optimize the dynamics loss~\eqref{eq:dynamics_loss} against this downsampled joint configuration trajectory.

\subsubsection{Baseline Methods}
We compare \chainsplat with {\it Particle-Grid Neural Dynamics} (PGND)~\cite{zhang2024particle} and {\it PhysTwin}~\cite{jiang2025phystwin}, two state-of-the-art algorithms for constructing DLO digital twins. 
PGND employs a hybrid Lagrangian--Eulerian neural dynamics model inspired by the Material Point Method (MPM), whereas PhysTwin uses a mass--spring model for physics simulation. 
Both methods utilize RGBD videos, reconstruct 3D particle trajectories using CoTracker~\cite{karaev2025cotracker3} and depth observations, and train their dynamics models using 3D particles as the state representation. 
To additionally model appearance, they construct Gaussians~\cite{kerbl20233d} based on the initial RGB frames and interpolate the motion of Gaussians using Linear Blend Skinning~\cite{sumner2007embedded} for rendering videos. 
We evaluate two variants of PhysTwin: {\it PhysTwin w/ 3D}, which uses a 3D reconstruction module~\cite{xiang2025structured} to fill the object's interior volume, and {\it PhysTwin w/o 3D}, which does not use this module.
We train all baselines using the code released with their respective papers. 
We only modify PhysTwin's action representation, as the original framework represents the action using a 3D point cloud of the human hand. Since our setup is based on a parallel-jaw gripper, we represent the action as a two-point point cloud sampled at the gripper fingertips. 
As with \chainsplat, we train a separate model for each trajectory, and evaluate each model by integrating the learned dynamics on the same base trajectory as for optimization.

\begin{figure*}[tbp]
    \centering
    \includegraphics[width=\linewidth]{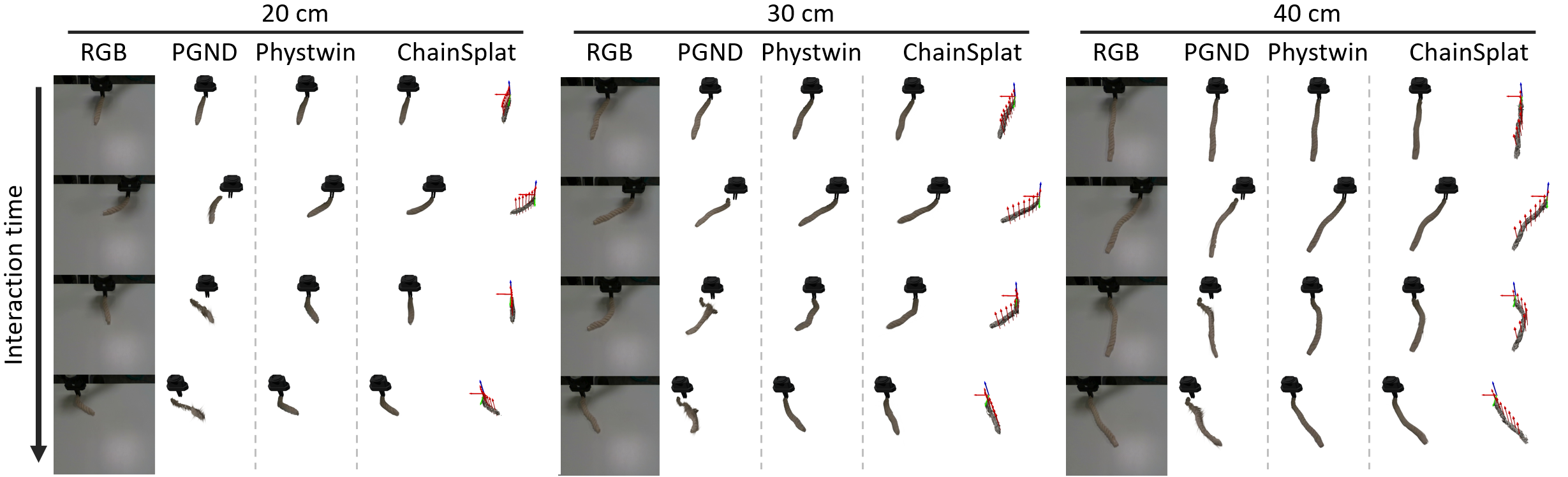}
    \vspace{-15pt}
    \caption{
    \chainsplat dynamics rollouts for three DLOs in tabletop scenarios, compared with PGND and PhysTwin w/o 3D.
    We visualize the ground-truth interaction video and RGB renderings of the dynamics rollout from each approach.
    Snapshots are shown at $t \in \{0.1, 0.7, 2.1, 3.5\},\text{s}$ for the $20$~cm DLO, $t \in \{0.1, 0.4, 1.3, 1.7\},\text{s}$ for the $30$~cm DLO, and $t \in \{0.1, 0.7, 1.9, 3.0\},\text{s}$ for the $40$~cm DLO.
    For intuitive comparison, we overlay a gripper mesh on each rendered image based on the end-effector pose at the corresponding timestep.
    For \chainsplat, we additionally visualize the screw axes and 3D Gaussians transformed according to the predicted joint configuration.
    }
    \vspace{-5pt}
    \label{fig:dynamics_learning_tabletop}
\end{figure*}

\begin{table*}
\centering
\caption{Accuracy of DLO dynamics learning, 3D geometry reconstruction, and RGB appearance for three DLOs in tabletop scenarios.}
\label{table:dynamics_learning_tabletop}
\renewcommand{\arraystretch}{1.2}
\setlength{\tabcolsep}{2.5pt}
\resizebox{\linewidth}{!}{
\begin{tabular}{lccccccccccccccccc}
& \multicolumn{5}{c}{$20$~cm} & & \multicolumn{5}{c}{$30$~cm} & & \multicolumn{5}{c}{$40$~cm} \\
\cline{2-6}
\cline{8-12}
\cline{14-18}

\multicolumn{1}{c}{\bf METHOD} & CD $\downarrow$ & IoU $\uparrow$ & PSNR $\uparrow$ & SSIM $\uparrow$ & LPIPS $\downarrow$ & & CD $\downarrow$ & IoU $\uparrow$ & PSNR $\uparrow$ & SSIM $\uparrow$ & LPIPS $\downarrow$ & & CD $\downarrow$ & IoU $\uparrow$ & PSNR $\uparrow$ & SSIM $\uparrow$ & LPIPS $\downarrow$ \\
\hline

PGND~\cite{zhang2024particle} & $0.0348$ & $0.402$ & $25.11$ & $0.986$ & $0.0171$ & & $0.0261$ & $0.511$ & $24.85$ & $0.982$ & $0.0212$ & & $0.0302$ & $0.466$ & $22.90$ & $0.973$ & $0.0315$ \\
Phystwin w/o 3D~\cite{jiang2025phystwin} & $\bm{0.0201}$ & $0.578$ & $27.58$ & $0.990$ & $0.0093$ & & $\bm{0.0171}$ & $0.636$ & $26.84$ & $0.985$ & $0.0133$ & & $\bm{0.0212}$ & $\bm{0.569}$ & $24.37$ & $0.978$ & $\bm{0.0193}$ \\
Phystwin w/ 3D~\cite{jiang2025phystwin} & $0.0301$ & $0.434$ & $25.78$ & $0.988$ & $0.0123$ & & $0.0240$ & $0.520$ & $25.23$ & $0.983$ & $0.0159$ & & $0.0241$ & $0.528$ & $24.04$ & $0.977$ & $0.0204$ \\
\chainsplat (ours) & $0.0263$ & $\bm{0.616}$ & $\bm{29.54}$ & $\bm{0.992}$ & $\bm{0.0083}$ & & $0.0220$ & $\bm{0.661}$ & $\bm{28.69}$ & $\bm{0.988}$ & $\bm{0.0114}$ & & $0.0333$ & $0.503$ & $\bm{24.61}$ & $\bm{0.980}$ & $0.0215$ \\
\hline
\end{tabular}
}
\end{table*}

\subsection{Kinematics-aware Geometry Recognition and Dynamics Learning}
\label{subsec:experiments-dynamics-learning}
In this section, we evaluate the performance of \chainsplat to learn the 3D geometry, appearance, kinematics, and dynamics of DLOs, demonstrating its advantages over existing state-of-the-art methods in DLO digital twin modeling. 
We use five evaluation metrics: the Chamfer Distance (CD) on 3D points obtained from depth images, the Intersection over Union (IoU) on object masks, and three RGB appearance metrics: Peak Signal-to-Noise Ratio (PSNR), Structural Similarity Index Measure (SSIM), and Learned Perceptual Image Patch Similarity (LPIPS)~\cite{zhang2018unreasonable}. 
All metrics evaluate the accuracy of the learned DLO dynamics by measuring how faithfully the predicted DLO state matches the ground truth over time, with CD and IoU placing particular emphasis on 3D geometry accuracy and PSNR, SSIM, and LPIPS focusing more on appearance fidelity. 
Since \chainsplat and the baselines are based on Gaussian splatting, we render depth and mask images as in~\cite{ye2025gsplat}. 
We compute the CD between the 3D point clouds back-projected from the rendered and ground-truth depth images, and the IoU between the rendered and ground-truth masks. 
The remaining three metrics (PSNR, SSIM, and LPIPS) are computed between the rendered and ground-truth RGB images. 
All metrics are averaged across all three camera views and all frames.

We compare all methods qualitatively and quantitatively in the tabletop and free-space experimental scenarios. We first examine the tabletop scenario. 
Figure~\ref{fig:dynamics_learning_tabletop} shows representative dynamics rollout results for the three DLOs for \chainsplat, PGND, and PhysTwin w/o 3D. 
\chainsplat generally demonstrates the best dynamics prediction and RGB appearance fidelity.
PGND produces reasonable predictions during the early stages of the interaction, but increasingly deviates from the ground truth over time, eventually failing to accurately capture even the motion of the grasped end. 
PhysTwin outperforms PGND and produces rollouts that more closely follow the ground-truth behavior, in some cases exhibiting qualitatively comparable performance to \chainsplat. 
Table~\ref{table:dynamics_learning_tabletop} presents the quantitative results for the tabletop scenario. 
Overall, \chainsplat outperforms the baselines across most metrics, displaying particularly strong performance in RGB appearance fidelity. We also observe that \chainsplat generally performs better as the DLO length decreases. 
We attribute this trend to the use of a fixed number of $8$ screw axes across all DLOs, which provides a finer spatial resolution for shorter DLOs thus enabling more accurate modeling of their deformations. Moreover, PGND consistently exhibits the lowest performance across all DLOs. 
PhysTwin improves upon PGND and achieves slightly lower CD than \chainsplat. 
However, it is worth noting that \chainsplat does not use ground-truth depth observations during dynamics learning, yet yields CD comparable to PhysTwin which explicitly leverages depth observations.
We also note that incorporating the 3D reconstruction module does not consistently improve the performance of PhysTwin, which we attribute to inaccurate 3D reconstructions resulting from noisy depth observations.

\begin{figure*}[tbp]
    \centering
    \includegraphics[width=\linewidth]{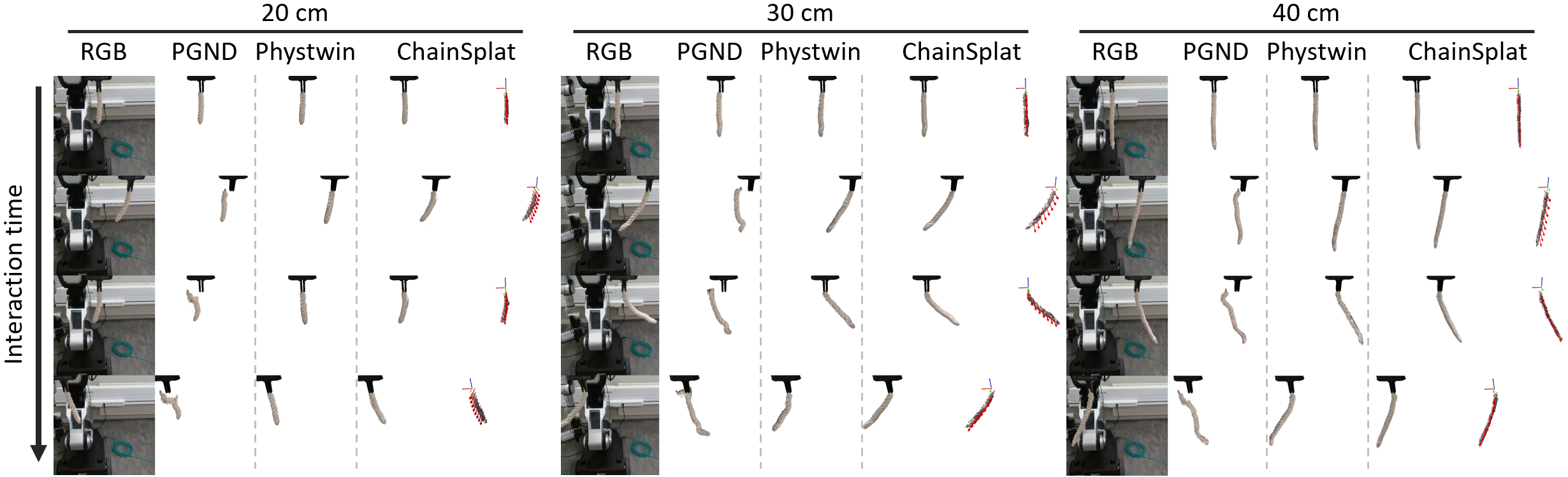}
    \vspace{-15pt}
    \caption{\chainsplat dynamics rollout results for three DLOs in free-space scenarios, compared with PGND and  PhysTwin w/o 3D.
    We visualize the ground-truth interaction video and RGB renderings of the dynamics rollout from each approach. 
    Snapshots are shown at $t \in \{0.1, 0.8, 2.0, 2.8\}\,\text{s}$ for the $20$~cm DLO, $t \in \{0.1, 1.0, 1.4, 1.7\}\,\text{s}$ for the $30$~cm DLO, and $t \in \{0.1, 0.4, 1.6, 2.2\}\,\text{s}$ for the $40$~cm DLO.
    For intuitive comparison, we overlay a gripper mesh on each rendered image based on the end-effector pose at the corresponding timestep.
    For \chainsplat, we additionally visualize the screw axes and 3D Gaussians transformed according to the predicted joint configuration.
    }
    \vspace{-5pt}
    \label{fig:dynamics_learning_freespace}
\end{figure*}

\begin{table*}
\centering
\caption{Accuracy of DLO dynamics learning, 3D geometry reconstruction, and RGB appearance for three DLOs in free-space scenarios.}
\label{table:dynamics_learning_freespace}
\renewcommand{\arraystretch}{1.2}
\setlength{\tabcolsep}{2.5pt}
\resizebox{\linewidth}{!}{
\begin{tabular}{lccccccccccccccccc}
& \multicolumn{5}{c}{$20$~cm} & & \multicolumn{5}{c}{$30$~cm} & & \multicolumn{5}{c}{$40$~cm} \\
\cline{2-6}
\cline{8-12}
\cline{14-18}

\multicolumn{1}{c}{\bf METHOD} & CD $\downarrow$ & IoU $\uparrow$ & PSNR $\uparrow$ & SSIM $\uparrow$ & LPIPS $\downarrow$ & & CD $\downarrow$ & IoU $\uparrow$ & PSNR $\uparrow$ & SSIM $\uparrow$ & LPIPS $\downarrow$ & & CD $\downarrow$ & IoU $\uparrow$ & PSNR $\uparrow$ & SSIM $\uparrow$ & LPIPS $\downarrow$ \\
\hline

PGND~\cite{zhang2024particle} & $0.0680$ & $0.273$ & $28.06$ & $0.985$ & $0.0193$ & & $0.0722$ & $0.228$ & $26.21$ & $0.979$ & $0.0255$ & & $0.0852$ & $0.216$ & $25.52$ & $0.974$ & $0.0324$ \\
Phystwin w/o 3D~\cite{jiang2025phystwin} & $0.0347$ & $0.573$ & $31.12$ & $0.989$ & $0.0103$ & & $0.0329$ & $0.549$ & $29.28$ & $0.984$ & $0.0143$ & & $0.0446$ & $0.475$ & $27.85$ & $0.979$ & $0.0205$ \\
Phystwin w/ 3D~\cite{jiang2025phystwin} & $0.0380$ & $0.549$ & $30.96$ & $0.988$ & $0.0113$ & & $0.0440$ & $0.472$ & $28.48$ & $0.982$ & $0.0179$ & & $0.0435$ & $0.481$ & $28.06$ & $0.979$ & $0.0198$ \\
\chainsplat (ours) & $\bm{0.0293}$ & $\bm{0.704}$ & $\bm{33.65}$ & $\bm{0.992}$ & $\bm{0.0084}$ & & $\bm{0.0265}$ & $\bm{0.678}$ & $\bm{31.84}$ & $\bm{0.988}$ & $\bm{0.0115}$ & & $\bm{0.0329}$ & $\bm{0.670}$ & $\bm{31.08}$ & $\bm{0.985}$ & $\bm{0.0144}$ \\
\hline
\end{tabular}
}
\end{table*}

\begin{table}
\centering
\caption{Average training time (minutes) and inference time (seconds)}
\label{table:train_test_time}
\begin{tabular}{lcccc}
\multicolumn{1}{c}{\multirow{2}{*}{\bf METHOD}} & \multirow{2}{*}{Training $\downarrow$} & \multicolumn{3}{c}{Inference} \\
\cline{3-5}
 & & Rollout $\downarrow$ & Render $\downarrow$ & Total $\downarrow$ \\
\hline
PGND~\cite{zhang2024particle} & $332.0$ & $2.02$ & $3.88$ & $5.89$ \\
Phystwin w/o 3D~\cite{jiang2025phystwin} & $27.0$ & $1.60$ & $2.70$ & $4.31$ \\
Phystwin w/ 3D~\cite{jiang2025phystwin} & $29.4$ & $1.88$ & $6.56$ & $8.44$ \\
\chainsplat (ours) & $\bm{17.6}$ & $\bm{1.28}$ & $\bm{1.14}$ & $\bm{2.41}$ \\
\hline
\end{tabular}
\end{table}

Figure~\ref{fig:dynamics_learning_freespace} and Table~\ref{table:dynamics_learning_freespace} present the dynamics rollout results for the three DLOs across all methods in the free-space scenario. 
While the overall trends are similar to those observed in the tabletop scenario, \chainsplat exhibits a clearer performance advantage in the free-space scenario, outperforming all baselines across all metrics and DLOs.
As illustrated in Figure~\ref{fig:dynamics_learning_freespace}, \chainsplat accurately captures the gravity-driven behavior of the DLOs.
In contrast, PGND performs substantially worse than in the tabletop scenario, struggling to capture the dynamic behavior of DLOs under gravity. 
This performance gap highlights the importance of incorporating appropriate physics priors for modeling gravity-driven dynamics. 
PhysTwin performs substantially better than PGND, benefiting from its physics-based mass--spring dynamics, but still exhibits larger prediction errors than \chainsplat.
These results suggest that the open-chain dynamics model of \chainsplat~\eqref{eq:chainsplat_dynamics} provides an effective physics prior for modeling DLO dynamics in free space.

Next, we compare the computational efficiency of \chainsplat against the baselines. 
Table~\ref{table:train_test_time} reports the training and inference times for a single $5$-second trajectory from the dataset, measured on a single NVIDIA A100 GPU. 
\chainsplat achieves the shortest training time of $17.6$ minutes, consisting of $9.2$ minutes for kinematics-aware geometry recognition and $8.4$ minutes for dynamics identification. 
At inference time, \chainsplat is also faster than the baselines in both dynamics rollout (i.e., dynamics integration) and RGB rendering.
In particular, \chainsplat enables highly efficient rendering as it require no additional processing of the Gaussians (e.g., LBS), prior to rendering. 
Overall, our results demonstrate that \chainsplat consistently outperforms state-of-the-art methods in constructing DLO digital twins in terms of both accuracy and computational efficiency.

\begin{figure}[!t]
    \centering
    \includegraphics[width=\linewidth]{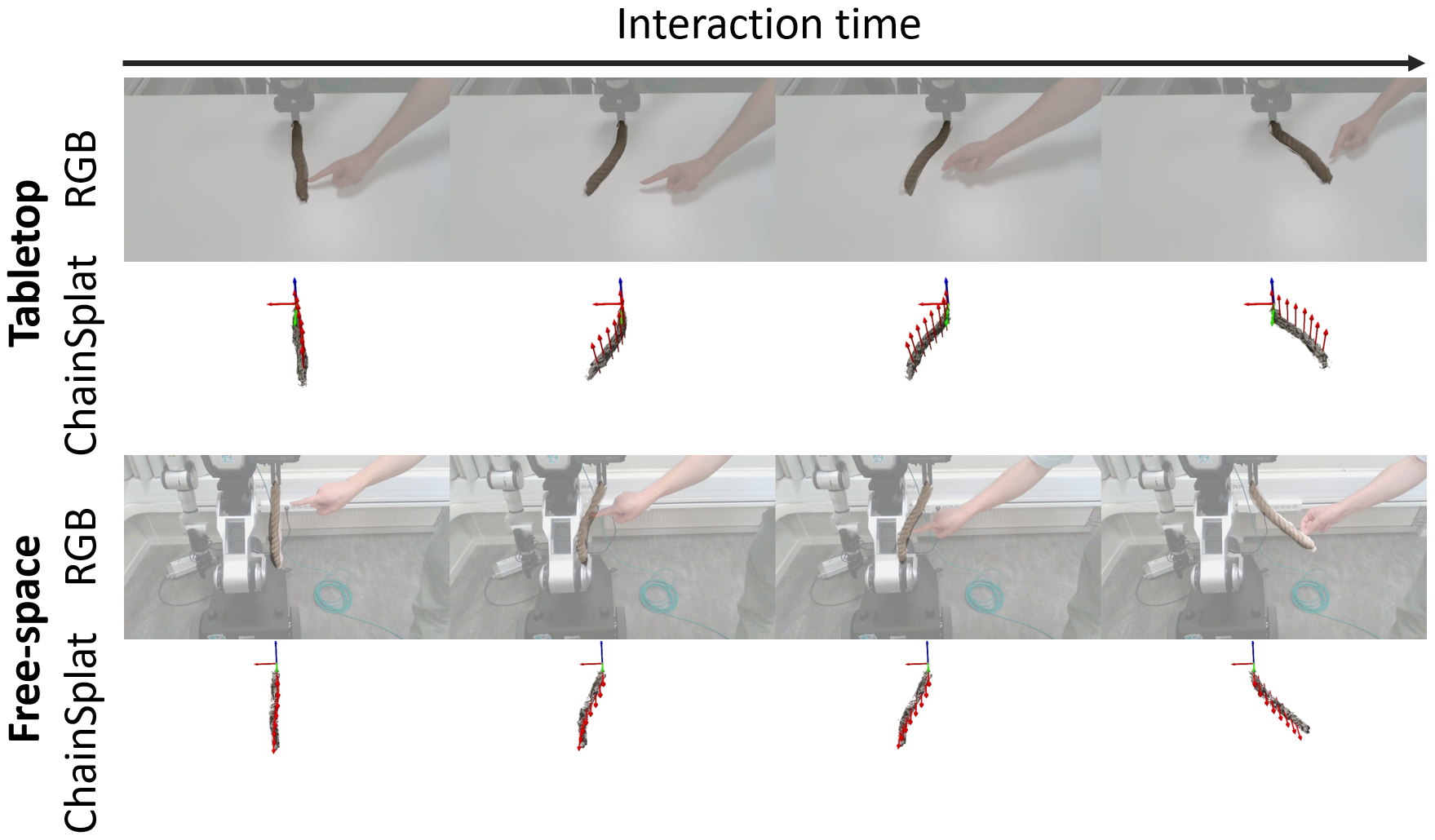}
    \vspace{-15pt}
    \caption{
    State estimation of a $30$cm-long rope in tabletop (snapshots at $t \in \{0.0, 4.1, 7.0, 13.7\}\,\text{s}$) and free-space scenarios (snapshots at $t \in \{0.0, 2.4, 5.4, 12.6\}\,\text{s}$). 
    Top row: observed RGB images from a single camera and overlay the RGB renderings from \chainsplat. 
    Bottom rows: screw axes and 3D Gaussians transformed according to the estimated joint configurations.
    }
    \vspace{-15pt}
    \label{fig:state_estimation_results}
\end{figure}

\subsection{State and External Force Estimation}
\label{subsec:experiments-estimation}
In this section, we demonstrate the effectiveness of \chainsplat for estimating DLO states and external forces from the stream of a single RGB camera. Specifically, we consider scenarios in which a human pushes and pulls the DLO either on a tabletop or in free space with the interaction recorded by a single RealSense RGB camera. We evaluate all three DLOs in both scenarios. For each DLO and scenario, we select a \chainsplat model that (1) achieves high performance according to the evaluation metrics reported in the previous section and (2) is trained on a trajectory exhibiting sufficiently rich DLO dynamics.

Figure~\ref{fig:state_estimation_results} shows the \chainsplat state estimation results in both tabletop and free-space scenarios. 
As a human manipulates the DLO, \chainsplat accurately estimates the corresponding joint configurations, resulting in rendered RGB images that closely match the observed images. 
On a single RTX $4090$ GPU, RGB observations are processed at approximately $5-6$ Hz, with most of the computation devoted to segmenting the DLO and human hand using SAM2~\cite{kirillov2023segment}. 
Note that, although the hand segmentation is not strictly required for state estimation, we subsequently use it to identify the contact point for contact force estimation. 
The state estimation itself runs at approximately $2-3$ Hz using $10$ iterations of the gradient-based optimization~\eqref{eq:state_estimation} on the same GPU.

\begin{figure}[!t]
    \centering
    \includegraphics[width=\linewidth]{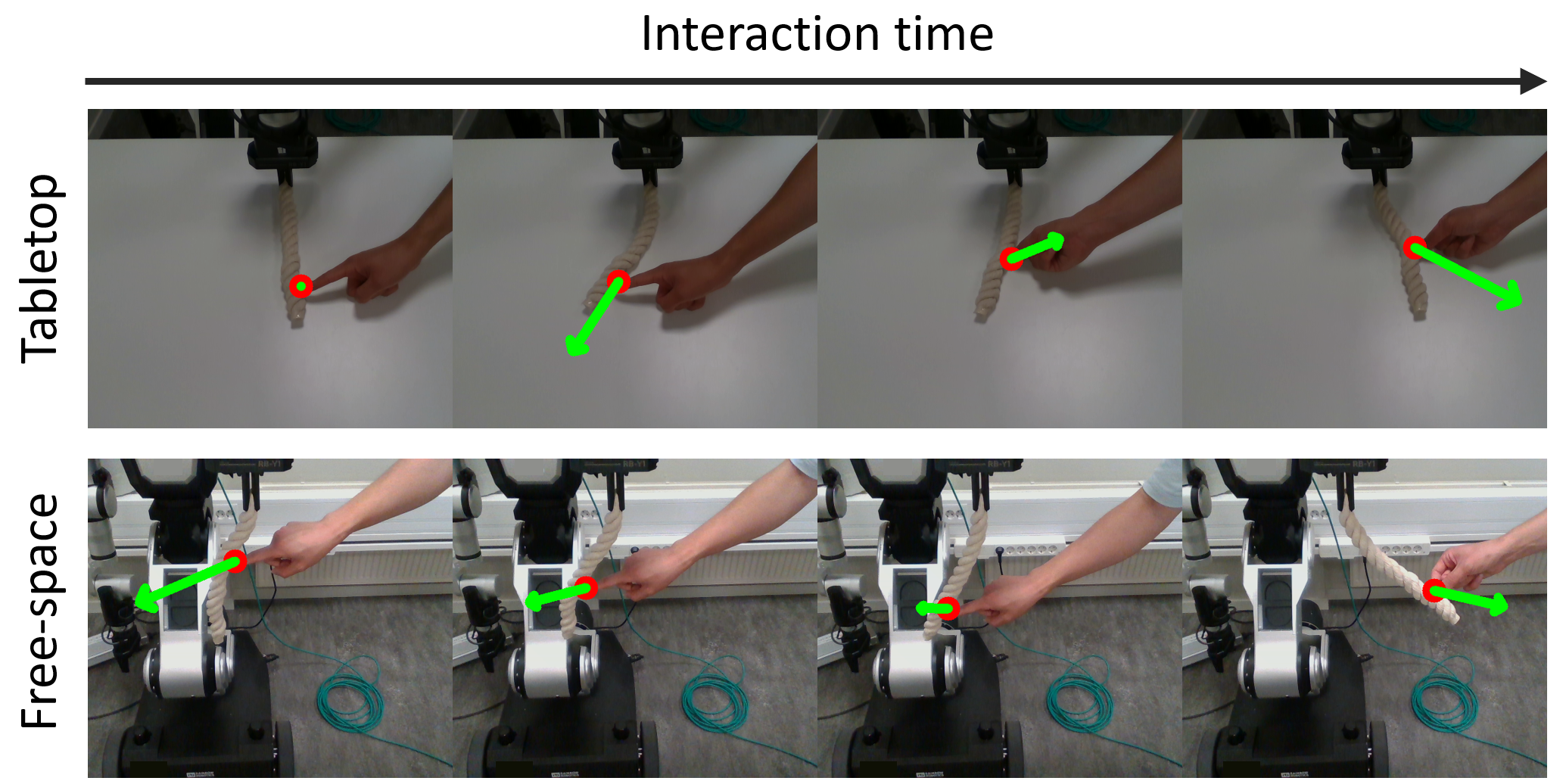}
    \vspace{-15pt}
    \caption{
    Force estimation during the interaction of a human with a $30$cm-long rope in tabletop (snapshots at $t \in \{0.0, 3.5, 8.8, 10.7\}\,\text{s}$) and free-space scenarios (snapshots at $t \in \{0.0, 3.5, 7.0, 10.8\}\,\text{s}$). 
    We visualize the observed RGB images and overlay the contact point (red circle) and the force applied to the DLO by a human hand (green arrow).
    }
    \vspace{-15pt}
    \label{fig:force_estimation_results}
\end{figure}

Using the states (i.e., joint configurations) estimated from the RGB stream, we compute the external forces applied to the DLO via~\eqref{eq:force_estimation_2}. 
The results are displayed in Figure~\ref{fig:force_estimation_results}. 
The force vectors are linearly normalized by the maximum force magnitude within each interaction video for better visualization. 
In the tabletop scenario, \chainsplat estimates contact forces consistent with the observed DLO motion while accounting for frictional forces between the DLO and the table (see Figure~\ref{fig:force_estimation_results}-top).
In the free-space scenario, the estimated contact forces appropriately counteract gravity to produce the observed DLO state (see Figure~\ref{fig:force_estimation_results}-bottom).
While the estimated forces are generally physically plausible, we observe that they can exhibit noise and substantial temporal variation. 
This is partly due to occasional inaccuracies and latency in state estimation caused by the limited number of optimization iterations used for real-time operation, as well as to the use of finite differences to estimate joint velocities which further amplifies noise.
Despite these limitations, our results demonstrate the potential of \chainsplat for state and external force estimation, providing a basis for future extensions to closed-loop control and contact-aware manipulation, respectively.

\subsection{Model-based Manipulation Results}
\label{subsec:experiment-manipulation}
Finally, we validate the proposed \chainsplat model-based manipulation framework on real-world DLO manipulation tasks. 
We consider both tabletop and free-space scenarios, with slightly different manipulation tasks and corresponding objective functions, as defined in~\eqref{eq:tabletop_objective}-\eqref{eq:freespace_objective} and discussed in Section~\ref{subsec:trajectory_optimization}. 
For tabletop manipulation, the goal is to bring an arbitrary point on the DLO into contact with the target point at the final timestep and stabilize the DLO at the resulting configuration. 
In contrast, for free-space manipulation, the goal is to bring the DLO tip into contact with the target point at any timestep during the trajectory.
We conduct three consecutive hitting trials for each DLO and scenario, with a different target point for each trial. 
We use a single RealSense camera to record the DLO during these experiments. 
Before each trial, we estimate the initial DLO state (i.e., joint configuration) from the initial RGB observation and use it to initialize the trajectory optimization.

\begin{figure*}[!t]
    \centering
    \includegraphics[width=\linewidth]{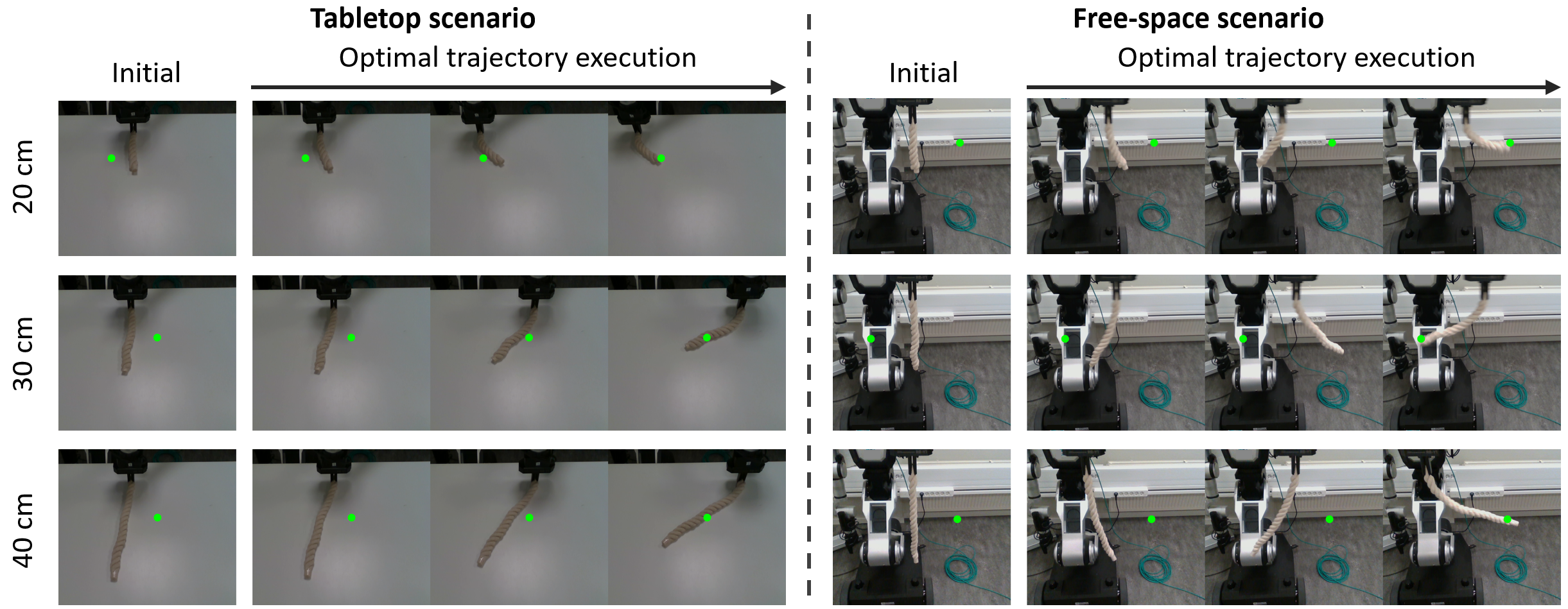}
    \vspace{-15pt}
    \caption{
    Model-based manipulation results for three DLOs in tabletop and free-space scenarios. 
    For each trial, we visualize the initial state of the DLO captured by a single RealSense camera together with the target goal points (green circles), as well snapshots of the robot executing the optimized trajectory and the resulting DLO motion toward the goal points.
    }
    \vspace{-5pt}
    \label{fig:manipulation_results}
\end{figure*}

Figure~\ref{fig:manipulation_results} shows representative robot and DLO trajectories obtained from \chainsplat model-based manipulation for all DLOs and scenarios. 
As shown in Figure~\ref{fig:manipulation_results}-left, in the tabletop scenario, \chainsplat successfully brings the DLOs into contact with the target goal points (green circles) and stabilizes them there. 
We emphasize that achieving this via open-loop trajectory optimization is nontrivial, as the optimizer must account for friction-induced changes in the DLO shape to accurately reach the target. 
These results demonstrate that \chainsplat effectively captures the DLO dynamics, including frictional effects, and that \chainsplat-based trajectory optimization successfully exploits these effects to derive effective model-based manipulation strategies.

Figure~\ref{fig:manipulation_results}-right displays representative robot and DLO trajectories obtained for the free-space scenario.
This task requires more dynamic DLO motions than the tabletop task, since the robot must exploit the transient, gravity-driven motion of the DLO to bring its tip into contact with the target, rather than stabilizing it there. 
We observe that \chainsplat successfully generates motions that bring the tip of the DLO into contact with the target by exploiting gravity-induced oscillations.
This demonstrate that \chainsplat effectively captures the gravity-driven dynamics of the DLO, which are then successfully exploited by the trajectory optimization for the hitting task.
Moreover, we observe that the optimized trajectories display different behaviors for DLOs of different lengths. 
For the shortest rope ($20$ cm), the optimization produces a spatially shorter but faster motion, whereas for the longest rope ($40$ cm), it produces a spatially larger but slower motion. 
This behavior is consistent with the physical tendency of longer DLOs to exhibit lower natural frequencies and longer characteristic oscillation periods, suggesting that \chainsplat's physics-based structure naturally reflects the underlying physical characteristics of the DLO in the optimized manipulation trajectories.

Table~\ref{table:manipulation} ablates the dynamics learning component of \chainsplat on the target-hitting accuracy. We compare \chainsplat with identified dynamics parameters with the variant {\it \chainsplat w/o dyn} which does not perform dynamics identification and instead uses the initial dynamics parameters. 
For the tabletop scenario, we measure the CD between the target point and the full DLO point cloud back-projected from depth images at the final timestep. 
For the free-space scenario, we extract the point cloud corresponding to the DLO tip from the depth images and report the minimum CD to the target point over all timesteps during execution.
As shown in Table~\ref{table:manipulation}, without dynamics identification, \chainsplat relies on inaccurate initial dynamics parameters, resulting in substantially degraded manipulation accuracy. 
In contrast, \chainsplat with identified dynamics parameters achieves high manipulation accuracy with an average target error of less than $1$ cm.
Overall, our results demonstrate that \chainsplat with identified dynamics parameters can be effectively used for accurate model-based DLO manipulation.

\begin{table}[t]
\centering
\caption{Accuracy of hitting target points (in cm) with DLOs using \chainsplat with initial and optimized dynamics parameters.}
\label{table:manipulation}
\resizebox{\linewidth}{!}{
\begin{tabular}{llccc}
& \multicolumn{1}{c}{\bf METHOD} & $20$~cm & $30$~cm & $40$~cm \\
\hline
\multirow{2}{*}{\rotatebox[origin=c]{90}{\scriptsize\textit{Table}}} & \chainsplat w/o dyn & $4.43\,{\scriptstyle\pm\,3.78}$ & $2.21\,{\scriptstyle\pm\,2.86}$ & $2.50\,{\scriptstyle\pm\,2.55}$ \\
 & \chainsplat & $\bm{0.08\,{\scriptstyle\pm\,0.02}}$ & $\bm{0.07\,{\scriptstyle\pm\,0.00}}$ & $\bm{0.46\,{\scriptstyle\pm\,0.66}}$ \\
\hline
\multirow{2}{*}{\rotatebox[origin=c]{90}{\scriptsize\textit{Free}}} & \chainsplat w/o dyn & $3.06\,{\scriptstyle\pm\,1.06}$ & $5.05\,{\scriptstyle\pm\,3.60}$ & $11.97\,{\scriptstyle\pm\,2.07}$ \\
 & \chainsplat & $\bm{1.50\,{\scriptstyle\pm\,0.29}}$ & $\bm{0.95\,{\scriptstyle\pm\,1.08}}$ & $\bm{0.22\,{\scriptstyle\pm\,0.10}}$ \\
\hline
\end{tabular}
}
\end{table}

%% file: 08discussion/v2.tex
\section{Discussions and Future Extensions}
Our experiments demonstrated that \chainsplat achieves state-of-the-art performance in DLO dynamics prediction, 3D geometry reconstruction, and RGB rendering in dynamic scenarios, while enabling real-time state and force estimation, as well as efficient and accurate model-based trajectory optimization for diverse DLO manipulation tasks.
In this section, we discuss potential extensions and applications of \chainsplat along three directions: 
(1) developing an end-to-end optimization framework beyond the current two-stage \chainsplat optimization procedure, 
(2) developing a single \chainsplat model that generalizes to diverse objects and interaction scenarios, and
(3) extending the proposed \chainsplat-based manipulation framework to closed-loop control. 
Together, these directions would advance \chainsplat toward more generalizable DLO digital twins and more versatile manipulation capabilities.

\subsubsection{End-to-End Optimization of \chainsplat}
The current optimization framework of \chainsplat follows a two-stage procedure consisting of kinematics-aware geometry recognition and dynamics identification. 
Decomposing these two problems can introduce systematic bias into dynamics learning as inaccurate estimates of kinematic parameters (e.g., screw axes and joint configurations) naturally lead to biased estimates of the dynamic parameters~\cite{kwon2021kinodynamic}. 
In other words, errors introduced during the kinematics-aware recognition stage cannot be corrected during the subsequent dynamics identification. 
Jointly learning a unified kinodynamic model could therefore improve the accuracy of the resulting dynamics representation by explicitly accounting for such coupled estimation errors.
Moreover, the dynamics of an open-chain system inherently depend not only on its dynamic parameters but also on its underlying kinematic structure, specifically on the screw parameters. 
An end-to-end optimization framework would thus enable the kinematic structure and dynamics parameters to be jointly optimized, potentially resolving ambiguities and redundant solutions in the kinematic representation that can arise when relying solely on kinematics-aware geometry recognition.

\subsubsection{Generalizable \chainsplat across Diverse Interactions and Scenarios}
A straightforward approach yo obtain a single generalizable \chainsplat model capable of handling diverse objects (e.g, DLOs of different lengths) and interaction scenarios would be to collect a large number of interaction trajectories to jointly train a single model. 
However, exhaustively collecting data covering all possible combinations of interaction patterns, environmental conditions, and DLO properties would require a prohibitively large number of real-world interaction trajectories.
Moreover, the choice of training trajectories can significantly affect the quality of the identified dynamics parameters. 
To address these limitations, \chainsplat could be extended via adaptive control, which would allow the dynamics parameters to be updated online during manipulation, thus compensating for modeling errors in a task-specific manner~\cite{lee2018natural}. 
In parallel, we could to reduce the amount of data required to obtain a generalizable \chainsplat model by employing optimal excitation trajectory design. This technique designs motions that make the dynamic system's underlying dynamic properties maximally informative for identification to generate a compact yet informative set of trajectories according to information-theoretic criteria~\cite{lee2021optimal}.
Such methodologies have been extensively studied for robotic systems, including open-chain manipulators, and could be naturally integrated into \chainsplat's open-chain dynamics formulation.
Combining online adaptation and excitation-aware data collection therefore represents a promising direction toward building more generalizable DLO digital twins.

\subsubsection{Closed-loop Reactive Manipulation of DLOs with \chainsplat}
The final aspect concerns leveraging \chainsplat for closed-loop manipulation. 
In practical DLO manipulation, target goal points may change during execution and unexpected obstacles, e.g., moving objects or humans, may enter the workspace. 
In such cases, it is essential that the robot reactively adapts its actions to changes in the environment. 
One possible approach is to learn a low-dimensional trajectory manifold offline from optimal trajectories collected for diverse target goal points and use it for rapid and reactive trajectory adaptation online~\cite{lee2024differentiable}. 
Alternatively, since \chainsplat provides a digital twin of the DLO, goal-conditioned policies that directly map the current state to robot actions~\cite{torne2024reconciling} could be trained in simulation and subsequently deployed in the real world.
Importantly, the low-dimensional state space of \chainsplat parameterized by joint configurations and the accompanying state estimator can facilitate efficient offline trajectory generation and policy learning, while enabling real-time state feedback for rapid closed-loop manipulation.

%% file: 09conclusion/v1.tex
\section{Conclusion}
This paper introduced \chainsplat, a physics-inspired model that jointly represents the 3D geometry, appearance, kinematics, and dynamics of deformable linear objects through a compact open-chain formulation. 
By combining analytic open-chain kinematics and dynamics based on screw theory with link-wise Gaussian splatting, \chainsplat provides both a low-dimensional state representation for efficient dynamics modeling and a high-fidelity representation of object geometry and appearance, operating directly on multi-view RGB videos without intermediate processing stages or auxiliary depth inputs. 
We developed a two-stage, fully-differentiable optimization framework that first jointly recovers the kinematics-aware geometry, appearance, and joint configuration trajectories of the DLO, and subsequently identifies its dynamic parameters. 
Our experiments demonstrated that \chainsplat achieves state-of-the-art performance in DLO digital-twin modeling, yielding accurate dynamics prediction, faithful 3D geometry reconstruction, and high-quality RGB rendering. 
Building upon the learned digital twin, we further demonstrated real-time state and external force estimation, as well as efficient and accurate gradient-based trajectory optimization for real-world DLO manipulation. 
Together, these results establish \chainsplat as a practical framework for learning physically-grounded DLO digital twins and for model-based robotic manipulation of DLOs.

%% file: references.bib
@string{RAM    		= "{IEEE} Robotics and Automation Magazine"}

@string{RAS    		= "Robotics and Autonomous Systems"}

@string{T-RO		= "{IEEE} Trans. on Robotics"}

@string{DSMC 		= "{ASME} Journal of Dynamic Systems, Measurement, and Control"}

@string{RAL 		= "{IEEE} Robotics and Automation Letters"}

@string{JMLR 		= "Journal of Machine Learning Research"}

@string{FRAI        = "Frontiers in Robotics and AI"}

@string{SCIR        = "Science Robotics"}

@string{AMR         = "Applied Mechanics Reviews"}

@string{TOG         = "{ACM} Transactions on Graphics"}

@string{DSMC        = "{ASME} Journal of Dynamic Systems, Measurement, and Control"}

@string{Automatica  = "Automatica"}

@string{IROS    	= "{IEEE/RSJ} Intl. Conf. on Intelligent Robots and Systems ({IROS})"}

@string{ICRA    	= "{IEEE} Intl. Conf. on Robotics and Automation ({ICRA})"}

@string{R:SS    	= "Robotics: Science and Systems ({R:SS})"}

@string{ICML   		= "Intl. Conf. on Machine Learning ({ICML})"}

@string{NeurIPS 		= "Neural Information Processing Systems ({NeurIPS})"}

@string{ICCV 		= "Intl. Conf. on Computer Vision ({ICCV})"}

@string{CVPR 		= "Conf. on Computer Vision and Pattern Recognition ({CVPR})"}

@string{CoRL 		= "Conference on Robot Learning ({CoRL})"}

@string{ICLR        = "Intl. Conf. on Learning Representations ({ICLR})"}

@string{ECCV        = "European Conference on Computer Vision ({ECCV})"}

@STRING{SIGGRAPH    = "Proc. of the ACM SIGGRAPH"}

@string{WAFR        = "International Workshop on the Algorithmic Foundations of Robotics ({WAFR})"}

@book{wanner1996solving,
  title={Solving ordinary differential equations II},
  author={Wanner, Gerhard and Hairer, Ernst},
  volume={375},
  year={1996},
  publisher={Springer Berlin Heidelberg New York}
}

@article{arriola2020modeling,
  title={Modeling of deformable objects for robotic manipulation: A tutorial and review},
  author={Arriola-Rios, Veronica E and Guler, Puren and Ficuciello, Fanny and Kragic, Danica and Siciliano, Bruno and Wyatt, Jeremy L},
  journal=FRAI,
  volume={7},
  year={2020},
  publisher={Frontiers Media SA}
}

@article{yin2021modeling,
  title={Modeling, learning, perception, and control methods for deformable object manipulation},
  author={Yin, Hang and Varava, Anastasia and Kragic, Danica},
  journal=SCIR,
  volume={6},
  number={54},
  year={2021},
  publisher={American Association for the Advancement of Science}
}

@article{ai2025review,
  title={A review of learning-based dynamics models for robotic manipulation},
  author={Ai, Bo and Tian, Stephen and Shi, Haochen and Wang, Yixuan and Pfaff, Tobias and Tan, Cheston and Christensen, Henrik I and Su, Hao and Wu, Jiajun and Li, Yunzhu},
  journal=SCIR,
  volume={10},
  number={106},
  year={2025},
  publisher={American Association for the Advancement of Science}
}

@article{dinkel2025dlo,
  title={{DLO}-splatting: Tracking deformable linear objects using 3d {G}aussian splatting},
  author={Dinkel, Holly and B{\"u}sching, Marcel and Longhini, Alberta and Coltin, Brian and Smith, Trey and Kragic, Danica and Bj{\"o}rkman, M{\aa}rten and Bretl, Timothy},
  journal={arXiv preprint arXiv:2505.08644},
  year={2025}
}

@incollection{sumner2007embedded,
  title={Embedded deformation for shape manipulation},
  author={Sumner, Robert W and Schmid, Johannes and Pauly, Mark},
  booktitle=SIGGRAPH,
  pages={80--es},
  year={2007}
}

@inproceedings{torne2024reconciling,
  title={Reconciling reality through simulation: A real-to-sim-to-real approach for robust manipulation},
  author={Torne, Marcel and Simeonov, Anthony and Li, Zechu and Chan, April and Chen, Tao and Gupta, Abhishek and Agrawal, Pulkit},
  booktitle=R:SS,
  year={2024}
}

@inproceedings{nair2017combining,
  title={Combining self-supervised learning and imitation for vision-based rope manipulation},
  author={Nair, Ashvin and Chen, Dian and Agrawal, Pulkit and Isola, Phillip and Abbeel, Pieter and Malik, Jitendra and Levine, Sergey},
  booktitle=ICRA,
  pages={2146--2153},
  year={2017},
}

@inproceedings{yan2021learning,
  title={Learning predictive representations for deformable objects using contrastive estimation},
  author={Yan, Wilson and Vangipuram, Ashwin and Abbeel, Pieter and Pinto, Lerrel},
  booktitle=CoRL,
  pages={564--574},
  year={2021},
  organization={PMLR}
}

@inproceedings{ma2022learning,
  title={Learning latent graph dynamics for visual manipulation of deformable objects},
  author={Ma, Xiao and Hsu, David and Lee, Wee Sun},
  booktitle=ICRA,
  pages={8266--8273},
  year={2022},
}

@article{liu2023model,
  title={Model-based control with sparse neural dynamics},
  author={Liu, Ziang and Zhou, Genggeng and He, Jeff and Marcucci, Tobia and Li, Fei-Fei and Wu, Jiajun and Li, Yunzhu},
  journal=NeurIPS,
  pages={6280--6296},
  year={2023}
}

@inproceedings{driess2023learning,
  title={Learning multi-object dynamics with compositional neural radiance fields},
  author={Driess, Danny and Huang, Zhiao and Li, Yunzhu and Tedrake, Russ and Toussaint, Marc},
  booktitle=CoRL,
  pages={1755--1768},
  year={2023},
  organization={PMLR}
}

@inproceedings{zhang2024adaptigraph,
  title={AdaptiGraph: Material-Adaptive Graph-Based Neural Dynamics for Robotic Manipulation},
  author={Zhang, Kaifeng and Li, Baoyu and Hauser, Kris and Li, Yunzhu},
  booktitle=R:SS,
  year={2024}
}

@inproceedings{zhang2025dynamic,
  title={Dynamic 3D {G}aussian Tracking for Graph-Based Neural Dynamics Modeling},
  author={Zhang, Mingtong and Zhang, Kaifeng and Li, Yunzhu},
  booktitle=CoRL,
  pages={1851--1862},
  year={2025},
  organization={PMLR}
}

@inproceedings{huang2025particleformer,
  title={ParticleFormer: A 3D Point Cloud World Model for Multi-Object, Multi-Material Robotic Manipulation},
  author={Huang, Suning and Chen, Qianzhong and Zhang, Xiaohan and Sun, Jiankai and Schwager, Mac},
  booktitle=CoRL,
  pages={4941--4957},
  year={2025},
  organization={PMLR}
}

@inproceedings{duisterhof2024deformgs,
  title={Deformgs: Scene flow in highly deformable scenes for deformable object manipulation},
  author={Duisterhof, Bardienus P and Zhao, Mandi and Yao, Yunchao and Liu, Jia-Wei and Seidenschwarz, Jenny and Shou, Mike Zheng and Ramanan, Deva and Song, Shuran and Birchfield, Stan and Wen, Bowen and others},
  booktitle=WAFR,
  pages={263--282},
  year={2024},
  organization={Springer}
}

@article{yan2020self,
  title={Self-supervised learning of state estimation for manipulating deformable linear objects},
  author={Yan, Mengyuan and Zhu, Yilin and Jin, Ning and Bohg, Jeannette},
  journal=RAL,
  volume={5},
  number={2},
  pages={2372--2379},
  year={2020},
}

@inproceedings{yang2021learning,
  title={Learning to propagate interaction effects for modeling deformable linear objects dynamics},
  author={Yang, Yuxuan and Stork, Johannes A and Stoyanov, Todor},
  booktitle=ICRA,
  pages={1950--1957},
  year={2021},
}

@article{yang2022learning,
  title={Learning differentiable dynamics models for shape control of deformable linear objects},
  author={Yang, Yuxuan and Stork, Johannes and Stoyanov, Todor},
  journal=RAS,
  volume={158},
  pages={104258},
  year={2022},
  publisher={Elsevier}
}

@article{yu2022global,
  title={Global model learning for large deformation control of elastic deformable linear objects: An efficient and adaptive approach},
  author={Yu, Mingrui and Lv, Kangchen and Zhong, Hanzhong and Song, Shiji and Li, Xiang},
  journal=T-RO,
  volume={39},
  number={1},
  pages={417--436},
  year={2022},
}

@article{caporali2024deformable,
  title={Deformable linear objects manipulation with online model parameters estimation},
  author={Caporali, Alessio and Kicki, Piotr and Galassi, Kevin and Zanella, Riccardo and Walas, Krzysztof and Palli, Gianluca},
  journal=RAL,
  volume={9},
  number={3},
  pages={2598--2605},
  year={2024},
}

@inproceedings{preiss2022tracking,
  title={Tracking fast trajectories with a deformable object using a learned model},
  author={Preiss, James A and Millard, David and Yao, Tao and Sukhatme, Gaurav S},
  booktitle=ICRA,
  pages={1351--1357},
  year={2022},
}

@inproceedings{zhang2021deformable,
  title={Deformable linear object prediction using locally linear latent dynamics},
  author={Zhang, Wenbo and Schmeckpeper, Karl and Chaudhari, Pratik and Daniilidis, Kostas},
  booktitle=ICRA,
  pages={13503--13509},
  year={2021},
}

@inproceedings{zhong2024reconstruction,
  title={Reconstruction and simulation of elastic objects with spring-mass 3d {G}aussians},
  author={Zhong, Licheng and Yu, Hong-Xing and Wu, Jiajun and Li, Yunzhu},
  booktitle=ECCV,
  pages={407--423},
  year={2024},
  organization={Springer}
}

@inproceedings{zhang2024particle,
  title={Particle-Grid Neural Dynamics for Learning Deformable Object Models from {RGB-D} Videos},
  author={Zhang, Kaifeng and Li, Baoyu and Hauser, Kris and Li, Yunzhu},
  booktitle=R:SS,
  year={2025}
}

@inproceedings{jiang2025phystwin,
  title={Phystwin: Physics-informed reconstruction and simulation of deformable objects from videos},
  author={Jiang, Hanxiao and Hsu, Hao-Yu and Zhang, Kaifeng and Yu, Hsin-Ni and Wang, Shenlong and Li, Yunzhu},
  booktitle=ICCV,
  pages={7219--7230},
  year={2025}
}

@article{chen2026empm,
  title={{EMPM}: Embodied {MPM} for Modeling and Simulation of Deformable Objects},
  author={Chen, Yunuo and Hu, Yafei and Sun, Lingfeng and Kusnur, Tushar and Herlant, Laura and Jiang, Chenfanfu},
  journal=RAL,
  volume={11},
  number={4},
  pages={4179--4186},
  year={2026}
}

@inproceedings{friedl2025riemannian,
  title={A {R}iemannian framework for learning reduced-order {L}agrangian dynamics},
  author={Friedl, Katharina and Jaquier, No{\'e}mie and Lundell, Jens and Asfour, Tamim and Kragic, Danica},
  booktitle=ICLR,
  pages={57775--57802},
  year={2025}
}

@inproceedings{friedl2025learning,
  title={Learning {H}amiltonian Dynamics at Scale: A Differential-Geometric Approach},
  author={Friedl, Katharina and Jaquier, No{\'e}mie and Liao, Alyx and Kragic, Danica},
  booktitle=ICML,
  year={2026}
}

@article{friedl2026reduced,
  title={Reduced-order Control and Geometric Structure of Learned {L}agrangian Latent Dynamics},
  author={Friedl, Katharina and Jaquier, No{\'e}mie and Kim, Seungyeon and Lundell, Jens and Kragic, Danica},
  journal={arXiv preprint arXiv:2602.08963},
  year={2026}
}

@inproceedings{mamedov2024pseudo,
  title={Pseudo-rigid body networks: learning interpretable deformable object dynamics from partial observations},
  author={Mamedov, Shamil and Geist, A Ren{\'e} and Swevers, Jan and Trimpe, Sebastian},
  booktitle=IROS,
  pages={9542--9548},
  year={2024},
}

@article{mamedov2025learning,
  title={Learning deformable linear object dynamics from a single trajectory},
  author={Mamedov, Shamil and Geist, A Rene and Viljoen, Ruan and Trimpe, Sebastian and Swevers, Jan},
  journal=RAL,
  volume={10},
  number={7},
  pages={7635--7642},
  year={2025},
}

@inproceedings{chen2025differentiable,
  title={Differentiable Discrete Elastic Rods for Real-Time Modeling of Deformable Linear Objects},
  author={Chen, Yizhou and Zhang, Yiting and Brei, Zachary and Zhang, Tiancheng and Chen, Yuzhen and Wu, Julie and Vasudevan, Ram},
  booktitle=CoRL,
  pages={2996--3014},
  year={2025},
  organization={PMLR}
}

@article{xiang2023trackdlo,
  title={Track{DLO}: Tracking deformable linear objects under occlusion with motion coherence},
  author={Xiang, Jingyi and Dinkel, Holly and Zhao, Harry and Gao, Naixiang and Coltin, Brian and Smith, Trey and Bretl, Timothy},
  journal=RAL,
  volume={8},
  number={10},
  pages={6179--6186},
  year={2023},
}

@book{lynch2017modern,
  title={Modern robotics},
  author={Lynch, Kevin M and Park, Frank C},
  year={2017},
  publisher={Cambridge University Press}
}

@article{park2018geometric,
  title={Geometric algorithms for robot dynamics: A tutorial review},
  author={Park, Frank C and Kim, Beobkyoon and Jang, Cheongjae and Hong, Jisoo},
  journal=AMR,
  volume={70},
  number={1},
  pages={010803},
  year={2018},
  publisher={American Society of Mechanical Engineers}
}

@article{kerbl20233d,
  title={3d {G}aussian splatting for real-time radiance field rendering},
  author={Kerbl, Bernhard and Kopanas, Georgios and Leimk{\"u}hler, Thomas and Drettakis, George},
  journal=TOG,
  volume={42},
  number={4},
  pages={1--14},
  year={2023}
}

@inproceedings{kim2025screwsplat,
  title={ScrewSplat: An End-to-End Method for Articulated Object Recognition},
  author={Kim, Seungyeon and Junsu, HA and Kim, Young Hun and Lee, Yonghyeon and Park, Frank C},
  booktitle=CoRL,
  pages={309--335},
  year={2025},
  organization={PMLR}
}

@inproceedings{karaev2025cotracker3,
  title={Cotracker3: Simpler and better point tracking by pseudo-labelling real videos},
  author={Karaev, Nikita and Makarov, Yuri and Wang, Jianyuan and Neverova, Natalia and Vedaldi, Andrea and Rupprecht, Christian},
  booktitle=ICCV,
  pages={6013--6022},
  year={2025}
}

@inproceedings{xiang2025structured,
  title={Structured 3d latents for scalable and versatile 3d generation},
  author={Xiang, Jianfeng and Lv, Zelong and Xu, Sicheng and Deng, Yu and Wang, Ruicheng and Zhang, Bowen and Chen, Dong and Tong, Xin and Yang, Jiaolong},
  booktitle=CVPR,
  pages={21469--21480},
  year={2025}
}

@inproceedings{bergou2008discrete,
  title={Discrete elastic rods},
  author={Bergou, Mikl{\'o}s and Wardetzky, Max and Robinson, Stephen and Audoly, Basile and Grinspun, Eitan},
  booktitle=SIGGRAPH,
  pages={1--12},
  year={2008}
}

@book{wittbrodt2006dynamics,
  title={Dynamics of flexible multibody systems: rigid finite element method},
  author={Wittbrodt, Edmund and Adamiec-W{\'o}jcik, Iwona and Wojciech, Stanisaw},
  year={2006},
  publisher={Springer}
}

@article{moberg2014modeling,
  title={Modeling and parameter estimation of robot manipulators using extended flexible joint models},
  author={Moberg, Stig and Wernholt, Erik and Hanssen, Sven and Brog{\aa}rdh, Torgny},
  journal=DSMC,
  volume={136},
  number={3},
  pages={031005},
  year={2014},
  publisher={American Society of Mechanical Engineers}
}

@inproceedings{kirillov2023segment,
  title={Segment Anything},
  author={Kirillov, Alexander and Mintun, Eric and Ravi, Nikhila and Mao, Hanzi and Rolland, Chloe and Gustafson, Laura and Xiao, Tete and Whitehead, Spencer and Berg, Alexander C and Lo, Wan-Yen and others},
  booktitle=ICCV,
  pages={3992--4003},
  year={2023}
}

@inproceedings{lee2024differentiable,
  title={Differentiable Motion Manifold Primitives for Reactive Motion Generation under Kinodynamic Constraints},
  author={Lee, Yonghyeon},
  booktitle=ICRA,
  year={2026}
}

@article{kwon2021kinodynamic,
  title={Kinodynamic model identification: A unified geometric approach},
  author={Kwon, Jaewoon and Choi, Keunjun and Park, Frank C},
  journal=T-RO,
  volume={37},
  number={4},
  pages={1100--1114},
  year={2021},
}

@inproceedings{lee2018natural,
  title={A natural adaptive control law for robot manipulators},
  author={Lee, Taeyoon and Kwon, Jaewoon and Park, Frank C},
  booktitle=IROS,
  pages={1--9},
  year={2018},
}

@article{lee2021optimal,
  title={Optimal excitation trajectories for mechanical systems identification},
  author={Lee, Taeyoon and Lee, Bryan D and Park, Frank C},
  journal=Automatica,
  volume={131},
  pages={109773},
  year={2021},
  publisher={Elsevier}
}

@inproceedings{zhang2018unreasonable,
  title={The unreasonable effectiveness of deep features as a perceptual metric},
  author={Zhang, Richard and Isola, Phillip and Efros, Alexei A and Shechtman, Eli and Wang, Oliver},
  booktitle=CVPR,
  pages={586--595},
  year={2018},
}

@article{ye2025gsplat,
  title={gsplat: An open-source library for Gaussian splatting},
  author={Ye, Vickie and Li, Ruilong and Kerr, Justin and Turkulainen, Matias and Yi, Brent and Pan, Zhuoyang and Seiskari, Otto and Ye, Jianbo and Hu, Jeffrey and Tancik, Matthew and Angjoo Kanazawa},
  journal=JMLR,
  volume={26},
  number={34},
  pages={1--17},
  year={2025}
}
